\documentclass[11pt, a4paper]{thuc3i}
\usepackage[numbers,sort&compress]{natbib}  
\setheadertext{A Survey of Video Reasoning with Large Multimodal Models}

\renewcommand{\today}{2025-10-06}

\usepackage[utf8]{inputenc} %
\usepackage[T1]{fontenc}    %
\usepackage{hyperref}       %
\usepackage{url}            %
\usepackage{tabularx}
\usepackage{makecell}
\usepackage{array,ragged2e,booktabs}
\newcolumntype{P}[1]{>{\RaggedRight\arraybackslash}p{#1}}
\usepackage{longtable}
\usepackage{longtable}
\usepackage{amsfonts}       %
\usepackage{amsthm}         %

\usepackage{nicefrac}       %
\usepackage{microtype}      %
\usepackage{xcolor}         %
\usepackage{xspace}
\newcommand{\eg}{e.g.\xspace}
\usepackage{algorithm}
\usepackage{algorithmicx}
\usepackage{algpseudocode}

\definecolor{darkblue}{rgb}{0, 0, 0.5}
\hypersetup{colorlinks=true, citecolor=darkblue, linkcolor=darkblue, urlcolor=darkblue}

\usepackage{xurl}

\usepackage{soul}
\usepackage{amsmath}
\usepackage{subcaption}
\usepackage{graphicx}
\usepackage{changes}
\usepackage{amsmath,mathtools}
\usepackage{changes}
\usepackage{pifont}
\usepackage{tocloft}

\usepackage{lipsum}
\usepackage{colortbl}
\usepackage{wrapfig}
\usepackage{xspace}
\usepackage{multirow}
\usepackage{enumitem}
\usepackage{cleveref}
\usepackage{pifont}
\usepackage{listings}
\usepackage{fontawesome5} 

\usepackage{wrapfig}
\usepackage{caption}

\usepackage{makecell}

\usepackage{tabularx}
\usepackage{afterpage}

\usepackage[edges]{forest}

\usepackage[normalem]{ulem}
\usepackage{caption}
\usepackage{CJKutf8}
\usepackage{bbding}
\usepackage[most,skins,theorems]{tcolorbox}

\usepackage[tikz]{bclogo}
\usepackage[framemethod=tikz]{mdframed}
\definecolor{bgblue}{RGB}{245,243,253}
\definecolor{ttblue}{RGB}{91,194,224}
\definecolor{SFTColor}{RGB}{225,240,255}
\definecolor{RLColor}{RGB}{255,236,207}
\definecolor{BenchColor}{RGB}{255,228,235}

\mdfdefinestyle{mystyle}{%
  rightline=true,
  innerleftmargin=10,
  innerrightmargin=10,
  outerlinewidth=3pt,
  topline=false,
  rightline=true,
  bottomline=false,
  skipabove=\topsep,
  skipbelow=\topsep
}

\newtcolorbox{myboxi}[1][]{
  breakable,
  title=#1,
  colback=red!5,
  colbacktitle=red!5,
  coltitle=black,
  fonttitle=\bfseries,
  bottomrule=0pt,
  toprule=0pt,
  leftrule=2pt,
  rightrule=2pt,
  titlerule=0pt,
  arc=0pt,
  outer arc=0pt,
  colframe=red,
}

\newtcolorbox{myboxnote}[1][]{
  breakable,
  title=#1,
  colback=orange!0,
  colbacktitle=orange!0,
  coltitle=black,
  fonttitle=\bfseries,
  bottomrule=0pt,
  toprule=0pt,
  leftrule=2pt,
  rightrule=2pt,
  titlerule=0pt,
  arc=0pt,
  outer arc=0pt,
  colframe=orange,
}

\usepackage{epigraph}

\newtcolorbox{myboxii}[1][]{
  breakable,
  freelance,
  title=#1,
  colback=white,
  colbacktitle=white,
  coltitle=black,
  fonttitle=\bfseries,
  bottomrule=0pt,
  boxrule=0pt,
  colframe=white,
  overlay unbroken and first={
  \draw[red!75!black,line width=3pt]
    ([xshift=5pt]frame.north west) -- 
    (frame.north west) -- 
    (frame.south west);
  \draw[red!75!black,line width=3pt]
    ([xshift=-5pt]frame.north east) -- 
    (frame.north east) -- 
    (frame.south east);
  },
  overlay unbroken app={
  \draw[red!75!black,line width=3pt,line cap=rect]
    (frame.south west) -- 
    ([xshift=5pt]frame.south west);
  \draw[red!75!black,line width=3pt,line cap=rect]
    (frame.south east) -- 
    ([xshift=-5pt]frame.south east);
  },
  overlay middle and last={
  \draw[red!75!black,line width=3pt]
    (frame.north west) -- 
    (frame.south west);
  \draw[red!75!black,line width=3pt]
    (frame.north east) -- 
    (frame.south east);
  },
  overlay last app={
  \draw[red!75!black,line width=3pt,line cap=rect]
    (frame.south west) --
    ([xshift=5pt]frame.south west);
  \draw[red!75!black,line width=3pt,line cap=rect]
    (frame.south east) --
    ([xshift=-5pt]frame.south east);
  },
}

\definecolor{myblue}{rgb}{0.9, 0.1, 0.94}
\definecolor{mygreen}{rgb}{0.64, 0.56, 0.88}
\definecolor{myyellow}{rgb}{0.68, 0.6, 0.1}
\definecolor{fancygreen}{rgb}{0.33, 0.68, 0.20}
\definecolor{salmon}{rgb}{0.94, 0.52, 0.49}
\definecolor{tablegreen}{rgb}{0.82, 0.94, 0.75}
\definecolor{tableblue}{rgb}{0.81, 0.90, 0.94}
\definecolor{tablered}{rgb}{0.97, 0.85, 0.85}
\definecolor{tableorange}{rgb}{0.96, 0.85, 0.81}

\definecolor{myorange}{rgb}{1.0, 0.49, 0.0}	
\definecolor{tlgreen}{rgb}{0.33, 0.68, 0.20}

\newenvironment{itemize*}%
 {\leftmargini=10pt\begin{itemize}%
  \setlength{\itemsep}{0pt}%
  \setlength{\parskip}{0pt}%
  }%
 {\end{itemize}}
\newenvironment{enumerate*}%
 {\begin{enumerate}%
  \setlength{\itemsep}{0pt}%
  \setlength{\parskip}{0pt}}%
 {\end{enumerate}}

\tikzset{%
    every node/.style={font=\tiny},
    parent/.style =          {align=center,text width=2cm,rounded corners=3pt, line width=0.3mm, fill=gray!10,draw=gray!80},
    child/.style =           {align=center,text width=2.0cm,rounded corners=3pt, fill=blue!10,draw=blue!80,line width=0.3mm},
    grandchild/.style =      {align=center,text width=2cm,rounded corners=3pt},
    greatgrandchild/.style = {align=center,text width=1.5cm,rounded corners=3pt},
    greatgrandchild2/.style = {align=center,text width=1.5cm,rounded corners=3pt},    
    referenceblock/.style =  {align=center,text width=1.5cm,rounded corners=2pt},
    pretrain/.style =           {align=center,text width=2.0cm,rounded corners=3pt, fill=blue!10,draw=blue!80,line width=0.3mm},   
    pretrain_work/.style =           {align=center, text width=8.5cm,rounded corners=3pt, fill=blue!10,draw=blue!0,line width=0.3mm},  
    template/.style =           {align=center,text width=2.0cm,rounded corners=3pt, fill=red!10,draw=red!80,line width=0.3mm},   
    template_work/.style =           {align=center,text width=8.5cm,rounded corners=3pt, fill=red!10,draw=red!0,line width=0.3mm},    
    answer/.style =           {align=center,text width=2.0cm,rounded corners=3pt, fill= cyan!10,draw= cyan!80,line width=0.3mm},   
    answer_work/.style =           {align=center,text width=8.5cm,rounded corners=3pt, fill= cyan!10,draw= cyan!0,line width=0.3mm},      
    multiple/.style =           {align=center,text width=2.0cm,rounded corners=3pt, fill= orange!10,draw= orange!80,line width=0.3mm},   
    multiple_work/.style =           {align=center,text width=8.5cm,rounded corners=3pt, fill= orange!10,draw= orange!0,line width=0.3mm},        
    tuning/.style =           {align=center,text width=2.0cm,rounded corners=3pt, fill= magenta!10,draw= magenta!80,line width=0.3mm},   
    tuning_work/.style =           {align=center,text width=8.5cm,rounded corners=3pt, fill= magenta!10,draw= magenta!0,line width=0.3mm},          
}

\usepackage{tikz}

\newcommand{\lstbg}[3][0pt]{{\fboxsep#1\colorbox{#2}{\strut #3}}}

\lstdefinelanguage{diff}{
  basicstyle=\ttfamily\small,
  morecomment=[f][\lstbg{red!20}]-,
  morecomment=[f][\lstbg{green!20}]+,
}
\lstdefinelanguage{diffpython}{
  language=diff,
  morekeywords={def, if, else, for, while, return, import, from, as, class, with, try, except, finally, raise, lambda, and, or, not, in, is, None, True, False},
  morecomment=[l]{\#},
  morestring=[b]",
  morestring=[b]',
}

\definecolor{darkgreen}{RGB}{50,100,0}
\definecolor{darkred}{RGB}{200, 0, 0}
\definecolor{lightblue}{RGB}{220,235,250}

\tcbset{
  takeawaysbox/.style={
    title=Takeaways,
    colback=lightblue!80,
    colframe=black,
    fonttitle=\bfseries\small,
    coltitle=white,
    colbacktitle=black,
    enhanced,
    attach boxed title to top left={xshift=2.5mm,yshift=-2.5mm},
    boxed title style={rounded corners, size=small, colframe=black, colback=black},
    width=\linewidth,
    arc=3.5mm
  }
}

\definecolor{darkgreen}{RGB}{50,100,0}
\definecolor{darkred}{RGB}{200, 0, 0}

\NewDocumentCommand{\kaiyan}
{ mO{} }{\textcolor{purple}{\textsuperscript{\textit{kaiyan}}\textsf{\textbf{\small[#1]}}}}
\NewDocumentCommand{\yuxin}
{ mO{} }{\textcolor{cyan}{\textsuperscript{\textit{yuxin}}\textsf{\textbf{\small[#1]}}}}
\NewDocumentCommand{\bx}
{ mO{} }{\textcolor{green}{\textsuperscript{\textit{bx}}\textsf{\textbf{\small[#1]}}}}
\NewDocumentCommand{\at}
{ mO{} }{\textcolor{red}{\textsuperscript{\textit{AT}}\textsf{\textbf{\small[#1]}}}}
\NewDocumentCommand{\re}
{ mO{} }{\textcolor{blue}{\textsuperscript{\textit{RE}}\textsf{\textbf{\small[#1]}}}}
\NewDocumentCommand{\ybsun}
{ mO{} }{\textcolor{magenta}{\textsuperscript{\textit{youbang}}\textsf{\textbf{\small[#1]}}}}
\NewDocumentCommand{\runze}
{ mO{} }{\textcolor{orange}{\textsuperscript{\textit{runze}}\textsf{\textbf{\small[#1]}}}}

\definecolor{darkgreen}{RGB}{0,100,0} 
\NewDocumentCommand{\add}
{ mO{} }{\textcolor{darkgreen}{\textsuperscript{\textit{Maybe Consider Discuss}}\textsf{\textbf{[#1]}}}}

\setlist[itemize]{leftmargin=20pt}

\usepackage[edges]{forest}
\definecolor{hidden-blue}{RGB}{194,232,247}
\definecolor{hidden-black}{RGB}{20,68,106}

\usepackage{tabularx, booktabs, xcolor, colortbl}
\usepackage{pifont}   %
\usepackage{xstring}  %

\usepackage{multirow}
\usepackage{hyperref}
\usepackage{fontawesome5}

\usepackage[export]{adjustbox}

\definecolor{yes}{HTML}{C6EFCE}      %
\definecolor{no}{HTML}{FFC7CE}       %
\definecolor{partial}{HTML}{FFEB9C}  %
\definecolor{external}{HTML}{D9E1F2} %
\definecolor{hdr}{HTML}{F2F2F2}

\usepackage{amssymb}
\newcommand{\cmark}{\textcolor{darkgreen}{\boldmath$\checkmark$}}
\newcommand{\xmark}{\textcolor{darkred}{\boldmath$\times$}}

\newcommand{\cellstatus}[1]{%
  \begingroup
  \StrTrim{#1}[\statusval]%
  \IfStrEq{\statusval}{Yes}{\cellcolor{yes}\cmark}{}%
  \IfStrEq{\statusval}{No}{\cellcolor{no}\xmark}{}%
  \IfBeginWith{\statusval}{Yes (}{\cellcolor{yes}\cmark~\textit{\statusval\unskip}}{}%
  \IfStrEq{\statusval}{Partial}{\cellcolor{partial}\textbf{Partial}}{}%
  \IfStrEq{\statusval}{External}{\cellcolor{external}\textbf{External}}{}%
  \endgroup
}

\newcommand{\faHuggingFace}{%
  \raisebox{-0.13em}{%
    \includegraphics[height=1em]{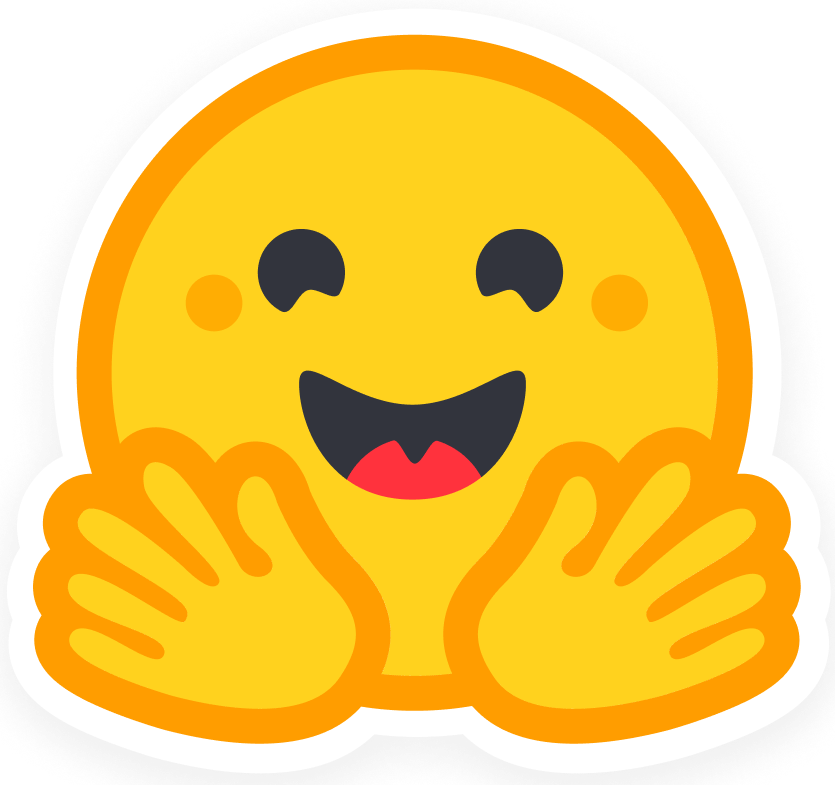}%
  }%
}

\newif\ifmaintitle
\maintitletrue
\title{\ifmaintitle\fontsize{19pt}{24pt}\selectfont\bfseries\fi Video-LMM Post-Training: A Deep Dive into Video Reasoning with Large~Multimodal~Models}

\author{%
    Yolo Yunlong Tang$^{1}$, Jing~Bi$^{1}$, Pinxin~Liu$^{1}$, Zhenyu~Pan$^{2}$, Zhangyun~Tan$^{1}$, Qianxiang Shen$^{1}$, Jiani~Liu$^{1}$, Hang~Hua$^{1}$, Junjia~Guo$^{1}$, Yunzhong~Xiao$^{3}$, Chao~Huang$^{1}$, Zhiyuan~Wang$^{4}$, Susan Liang$^{1}$, Xinyi~Liu$^{1}$, Yizhi~Song$^{5}$, Junhua~Huang$^{6}$, Jia-Xing~Zhong$^{7}$, Bozheng~Li$^{8}$, Daiqing~Qi$^{9}$, Ziyun~Zeng$^{1}$, Ali~Vosoughi$^{1}$, 
    Luchuan~Song$^{1}$, Zeliang~Zhang$^{1}$, Daiki~Shimada$^{10}$, Han~Liu$^{2}$, Jiebo~Luo$^{1}$, Chenliang~Xu$^{1}$
    \vspace{1mm} \\
    $^1$ University of Rochester \quad
    $^2$ Northwestern University \quad
    $^3$ CMU \quad
    $^4$ UCSB \quad
    $^5$ Purdue University \quad
    $^6$ UCLA \\
    $^7$ University of Oxford \quad
    $^8$ Brown University \quad
    $^9$ University of Virginia \quad
    $^{10}$ Sony Group Corporation 
    \vspace{1mm} \\
    \faEnvelope[regular]~\texttt{yunlong.tang@rochester.edu}  \quad
    \faGithub~\href{https://github.com/yunlong10/Awesome-Video-LMM-Post-Training}{yunlong10/Awesome-Video-LMM-Post-Training}
}

\begin{abstract}
    Video understanding represents the most challenging frontier in computer vision, requiring models to reason about complex spatiotemporal relationships, long-term dependencies, and multimodal evidence. The recent emergence of Video-Large Multimodal Models (Video-LMMs), which integrate visual encoders with powerful decoder-based language models, has demonstrated remarkable capabilities in video understanding tasks. However, the critical phase that transforms these models from basic perception systems into sophisticated reasoning engines—post-training—remains fragmented across the literature. This survey provides the first comprehensive examination of post-training methodologies for Video-LMMs, encompassing three fundamental pillars: supervised fine-tuning (SFT) with chain-of-thought, reinforcement learning (RL) from verifiable objectives, and test-time scaling (TTS) through enhanced inference computation. We present a structured taxonomy that clarifies the roles, interconnections, and video-specific adaptations of these techniques, addressing unique challenges such as temporal localization, spatiotemporal grounding, long video efficiency, and multimodal evidence integration. Through systematic analysis of representative methods, we synthesize key design principles, insights, and evaluation protocols while identifying critical open challenges in reward design, scalability, and cost-performance optimization. We further curate essential benchmarks, datasets, and metrics to facilitate rigorous assessment of post-training effectiveness. This survey aims to provide researchers and practitioners with a unified framework for advancing Video-LMM capabilities. Additional resources and updates are maintained at: \url{https://github.com/yunlong10/Awesome-Video-LMM-Post-Training}.
\end{abstract}

\begin{document}

\maketitle
\maintitlefalse 

\begin{figure}[!ht]
\centering
\includegraphics[width=\linewidth]{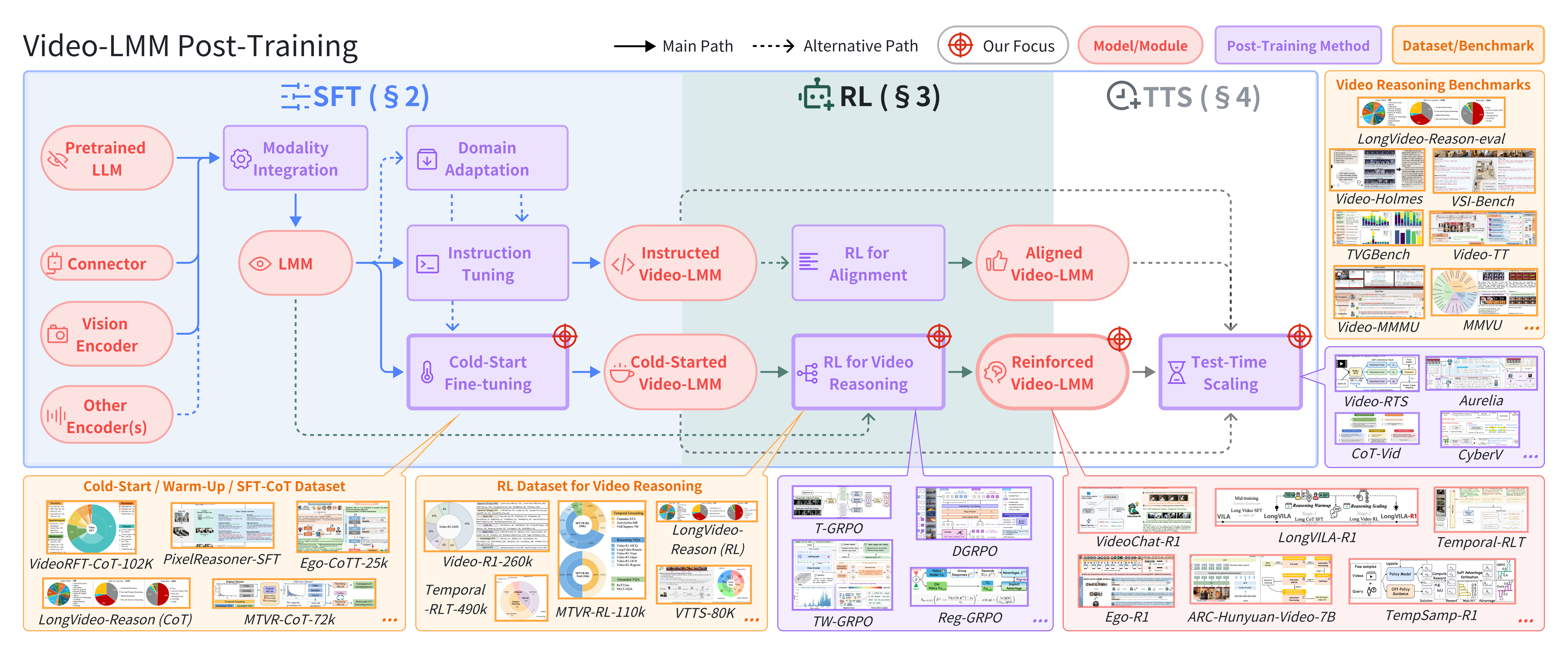}
\caption{
Overview of Video-LMM post-training and the scope of this survey.
}
\label{fig:teaser}
\end{figure}

\begin{figure}[!ht]
\centering
\includegraphics[width=\linewidth]{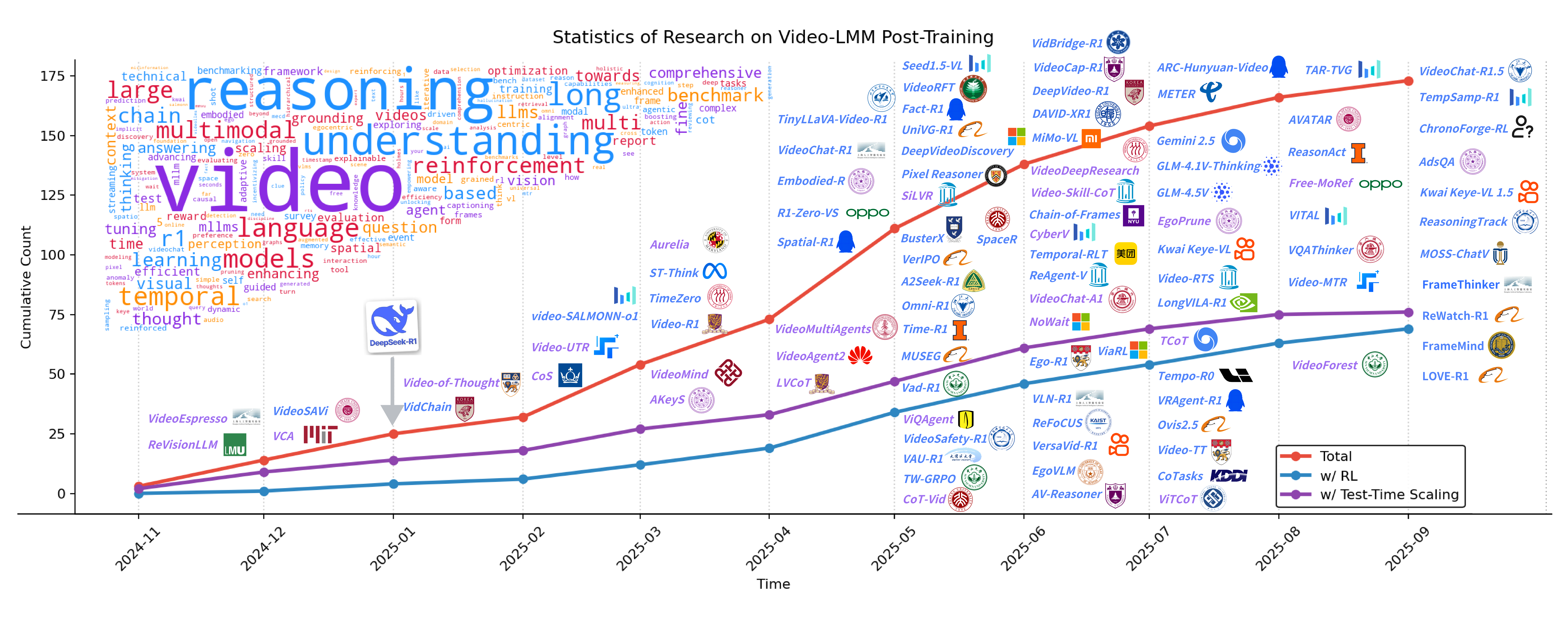}
\caption{
Research trends in Video-LMM post-training (November 2024 - September 2025). The word cloud is based on the titles of the papers.
}
\label{fig:timeline}
\end{figure}



\section{Introduction}

\begin{quote}
\raggedright
{\em One whale falls, ten thousand beings grow.}\\
\hfill-- {\scriptsize A modern saying, inspired by \emph{The Practice of the Wild}~\cite{snyder2020practice}}
\end{quote}

In recent years, Large Multimodal Models (LMMs)~\cite{comanici2025gemini,hurst2024gpt4o,team2023gemini,jaech2024openaigpto1,bai2025qwen25} have rapidly evolved from simple question-answering toward general problem-solving with interpretable long chain-of-thought (CoT) reasoning~\cite{wang2025multimodalchainofthoughtreasoningcomprehensive}. Video understanding, as one of the most comprehensive and challenging directions in computer vision, simultaneously involves complex spatiotemporal relationships, event causality, and long-term memory mechanisms, naturally demanding powerful language reasoning and task interface capabilities~\cite{vidllmsurvey}. Consequently, Video-LMMs featuring decoder-centric architectures have become the dominant paradigm~\cite{vidllmsurvey, zou2024secondshoursreviewingmultimodal}. These systems leverage strong LLMs as reasoning engines, employ video encoders to extract visual representations, align visual features to the LLM token embedding space through projection modules, and enable instruction understanding and answer generation, demonstrating superior initialization performance and generalization~\cite{vidllmsurvey}.

Video-language modeling has undergone three paradigm shifts: (1) the CNN+RNN era focused on temporal feature aggregation through recurrent architectures~\cite{yue2015beyond}; (2) Transformer-based video models, especially BERT-style/encoder-only joint representations, emphasized cross-modal alignment and retrieval through bidirectional encoding~\cite{sun2019videobert,neimark2021video}; (3) the current video encoder + decoder-based LLM architecture prioritizes the generality and composability of the language component while maximally reusing the knowledge and reasoning capabilities of pretrained LLMs~\cite{vidllmsurvey,hua2025finecaption, zou2024secondshoursreviewingmultimodal}. The key advantage lies in internet-scale self-supervised learning in the language domain, where next-token prediction enables knowledge, reasoning, and interface capabilities to emerge at scale under a unified objective. In contrast, the visual domain lacks an equivalent self-supervised learning method for efficiently processing internet-scale native video data. Although native multimodal approaches that jointly model vision and language end-to-end are being explored~\cite{coreteam2025mimovltechnicalreport, yang2025thinkingspacemultimodallarge}, they have yet to surpass the divide-and-conquer strategy in computational efficiency and engineering reusability.

Within this framework, \textbf{post-training} is the critical phase determining whether Video-LMMs progress from basic perception to sophisticated reasoning. As illustrated in \Cref{fig:teaser}, post-training encompasses three major components: (i) Supervised Fine-Tuning (SFT) incorporates CoT and reasoning style distillation to bootstrap reasoning formats and establish task-following behaviors~\cite{han2024videoespressolargescalechainofthoughtdataset, wang2025cotaskschainofthoughtbasedvideo, zhang2025vitcotvideotextinterleavedchainofthought}. (ii) Reinforcement Learning (RL) has evolved from RLHF, PPO, and DPO to R1-style/GRPO~\cite{guo2025deepseek} approaches that eliminate the need for preference data and explicit reward models, enabling enhanced reasoning and self-correction through verifiable objectives and systematic exploration~\cite{zhou2025reinforcedmllmsurveyrlbased, feng2025videor1reinforcingvideoreasoning, li2025videochatr1enhancingspatiotemporalperception, wang2025videortsrethinkingreinforcementlearning}. (iii) Test-Time Scaling (TTS) leverages increased inference computation for higher reliability through reasoning sample augmentation, voting mechanisms, self-consistency checks, external verifiers, and multi-path search~\cite{jin2025cotviddynamicchainofthoughtrouting, li2025veripocultivatinglongreasoning, chen2025expandingperformanceboundariesopensource}. This progression maintains close alignment with LLM community developments, offering transferability in theoretical principles and engineering practices.

Adapting these paradigms to video presents distinctive challenges differing substantially from static image-text scenarios. Temporal localization requires models to provide not only correct answers but also temporally precise responses anchored to specific segments~\cite{xu2025viarladaptivetemporalgrounding, luo2025musegreinforcingvideotemporal}. Spatiotemporal grounding demands consistency in tracking objects, parts, and actions across spatial and temporal dimensions~\cite{li2025videochatr1enhancingspatiotemporalperception, cheng2025vstarbenchmarkingvideollmsvideo}. Long video understanding necessitates sophisticated sampling strategies, adaptive routing, hierarchical viewing protocols, and effective caching~\cite{jiang2025stormtokenefficientlongvideo, hu2025mllmbasedvideoframe}. Multimodal evidence integration requires coordinated reasoning over video frames, textual captions, audio transcripts, and external knowledge~\cite{xiong2025streamingvideounderstandingmultiround, zhi2025videoagent2enhancingllmbasedagent, ghazanfari2025chainofframesadvancingvideounderstanding}. These characteristics have catalyzed video-specific post-training strategies: incorporating verifiable temporal and spatial rewards (tIoU, region consistency metrics) in RL frameworks; designing TTS methods that guide models to autonomously select informative frames and perform staged viewing with multi-round reflection and self-correction; and unifying diverse tasks (question answering, temporal localization, spatiotemporal grounding) within coherent alignment and optimization frameworks, establishing hierarchical pipelines for watching, thinking, locating, and answering~\cite{xu2025viarladaptivetemporalgrounding, luo2025musegreinforcingvideotemporal, jin2025cotviddynamicchainofthoughtrouting}.

Recent studies have successfully integrated GRPO/R1-style RL with extended reasoning TTS into video understanding, as illustrated in \Cref{fig:timeline}. Some emphasize verifiable reward design for temporal reasoning and localization~\cite{xu2025viarladaptivetemporalgrounding, luo2025musegreinforcingvideotemporal}, others extend to joint spatiotemporal grounding~\cite{li2025videochatr1enhancingspatiotemporalperception}, while others focus on long video scaling with efficient training and inference~\cite{chen2025scalingrllongvideos, jiang2025stormtokenefficientlongvideo}, and interactive viewing paradigms enabling thinking with video through evidence accumulation across iterations~\cite{fei2024videoofthoughtstepbystepvideoreasoning, jin2025cotviddynamicchainofthoughtrouting, zhang2025thinkingvideosmultimodaltoolaugmented}. This research wave has validated the feasibility of transferring LLM post-training paradigms to video understanding and revealed common challenges in data construction, reward robustness, evaluation protocol standardization, and cost-performance optimization, underscoring the need for a comprehensive survey examining Video-LMM reasoning methods from a post-training perspective.

In this survey, we focus on post-training for Video-LMMs, providing systematic coverage of key techniques across SFT, RL, and TTS, along with their specialized adaptations for video scenarios. We synthesize design principles and engineering insights from representative methods and discuss open challenges and future directions under unified evaluation and reporting standards. 

In short, the key contributions of this survey are as follows:
\begin{myboxi}[Contributions]
\begin{itemize}
    \item A comprehensive review of post-training methodologies for Video-LMMs, covering supervised fine-tuning, reinforcement learning, and test-time scaling as integral components of model optimization.
    \item A structured taxonomy of Video-LMM post-training techniques, clarifying their functional roles and interconnections, with insights into open challenges and future directions.
    \item Practical guidance introducing essential benchmarks, datasets, and evaluation metrics for assessing Video-LMM post-training effectiveness.
\end{itemize}
\end{myboxi}

\paragraph{Related Surveys.} Several surveys have reviewed video understanding with large language models \cite{vidllmsurvey,zhang2025videothinkingtestholistic, kumar2025videollmbenchmarksevaluationsurvey, zou2024secondshoursreviewingmultimodal,vtg_survey}, multimodal chain-of-thought reasoning \cite{wang2025multimodalchainofthoughtreasoningcomprehensive}, and reinforcement learning in LMMs \cite{zhou2025reinforcedmllmsurveyrlbased}. We also note the recent survey on reinforcement learning for large reasoning models \cite{zhang2025survey}, which provides broader context on RL-driven reasoning complementary to video post-training. While these works provide valuable perspectives on video-language modeling and reasoning techniques, our survey distinctly focuses on systematic organization and analysis of post-training methodologies specifically tailored for Video-LMMs, offering a unified treatment of SFT, RL, and TTS approaches.

\paragraph{Survey Structure.} \Cref{sec:sft} examines SFT for effective Video-LMM fine-tuning, especially CoT-SFT. \Cref{sec:rl} reviews LLM-based RL foundations before systematically analyzing RL algorithms, especially R1-style methods for video reasoning, including model configurations, data preparation, optimization strategies, and policy/reward design. \Cref{sec:tts} investigates video-specific TTS methods, emphasizing adaptive viewing mechanisms, multi-path reasoning strategies, and verification architectures. \Cref{sec:benchmarks} surveys datasets, benchmarks, and evaluation metrics. \Cref{sec:future} discusses future directions. Additional resources and updates are maintained at: \url{https://github.com/yunlong10/Awesome-Video-LMM-Post-Training}.

\section{Supervised Fine-Tuning for Video Reasoning}\label{sec:sft}
Supervised fine-tuning (SFT) serves as a pivotal stage that not only refines multimodal alignment, enhances instruction-following capability, and instills structured reasoning behaviors in Video-LMMs but also bridges large-scale pretraining and reinforcement learning (RL), laying the foundation for stable and generalizable video reasoning.

\begin{myboxi}[Takeaways]
\begin{itemize}
    \item Fixed-format CoT supervision enables imitation of reasoning patterns but provides limited flexibility for self-exploration and error correction compared to RL approaches, necessitating the transition to RL for learning abstract objectives and generalizing to complex, unseen scenarios.
    \item CoT-SFT has evolved from a standalone training paradigm to a critical cold-start phase for RL, providing structured reasoning formats (<think>, <answer>) and stable initialization that prevents instability in subsequent RL-driven policy optimization.

\end{itemize}
\end{myboxi}

\begin{figure}[!ht]
\footnotesize
\begin{forest}
    for tree={
        forked edges,
        grow'=0,
        draw,
        rounded corners,
        node options={align=center,},
        text width=2.7cm,
        s sep=6pt,
        calign=child edge, calign child=(n_children()+1)/2,
    },
    [\textbf{Video-LMM Post-Training}~\S~\ref{sec:sft}--\S~\ref{sec:benchmarks}, fill=gray!45, parent
        [Supervised Fine-Tuning (SFT)~\S~\ref{sec:sft}, for tree={fill=blue!45, answer}
            [Modality Integration, answer
                [{Video-LLaMA \cite{videollama}, LLaVA \cite{llava}, LLaVA-1.5 \cite{llava15}, LLaVA-OneVision~\cite{li2024llavaonevision}, PAVE \cite{liu2025pavepatchingadaptingvideo}}, answer_work]
            ]
            [Domain Adaptation, answer
                [{VTimeLLM \cite{huang2024vtimellm},  {HawkEye}~\cite{Wang2024HawkEye}, {VLM4HOI}~\cite{bansal2024hoiref},AVicuna \cite{tang2024avicuna}, {TimeChat}~\cite{Ren2024TimeChat}, Elysium \cite{wang2024elysium}}, answer_work]
            ]
            [Video Instruction Tuning, answer
                [{Video-LLaMA \cite{videollama}, Video-LLaVA~\cite{lin2023videollava}, Otter~\cite{li2023otter}, V2Xum-LLM~\cite{hua2024v2xum}, PLLaVA~\cite{xu2024pllava}, MMICT~\cite{Chen2023MMICT}, Artemis~\cite{qiu2024artemis}, AVicuna~\cite{tang2024avicuna}, VILA~\cite{Lin2024VILA}}, answer_work]
            ]
            [CoT Reasoning Fine-tuning, answer
                [{Video-of-Thought \cite{fei2024videoofthoughtstepbystepvideoreasoning}, VideoRFT \cite{wang2025videorftincentivizingvideoreasoning}, Video-R1 \cite{feng2025videor1reinforcingvideoreasoning}, CaRDiff \cite{tang2024cardiff}}, answer_work]
            ]
            [Video-grounded CoT Fine-tuning, answer
                [{VideoEspresso \cite{han2024videoespressolargescalechainofthoughtdataset}, ViTCoT \cite{zhang2025vitcotvideotextinterleavedchainofthought}, VideoCoT~\cite{wang2024videocot}, Video-CoT~\cite{zhang2025videocot}, CoTasks \cite{wang2025cotaskschainofthoughtbasedvideo}}, answer_work]
            ]
            [CoT-SFT Dataset, answer
                [{VideoRFT-CoT-102K \cite{wang2025videorftincentivizingvideoreasoning}, Ego-CoTT-25k \cite{tian2025egor1chainoftoolthoughtultralongegocentric}, LongVideo-Reason-CoT \cite{chen2025scalingrllongvideos}, TVG-Coldstart-13K~\cite{chen2025datasets}, MTVR-CoT-72k \cite{zhang2025thinkingvideosmultimodaltoolaugmented}, VideoEspresso \cite{han2024videoespressolargescalechainofthoughtdataset}, CoTasks \cite{wang2025cotaskschainofthoughtbasedvideo}, MECD \cite{chen2024mecdunlockingmultieventcausal}}, answer_work]
            ]
            [CoT-SFT Finetuned Video-LMMs, answer
                [{Eagle 2.5~\cite{chen2025eagle}, Video-R1~\cite{feng2025videor1reinforcingvideoreasoning}, CoS~\cite{hu2025coschainofshotpromptinglong},  DeepVideoDiscovery~\cite{zhang2025deepvideodiscoveryagentic}, EgoPrune~\cite{li2025egopruneefficienttokenpruning}, EgoVLM~\cite{vinod2025egovlmpolicyoptimizationegocentric}, Embodied-R~\cite{lin2025embrace3kembodiedreasoningaction}, MUSEG~\cite{luo2025musegreinforcingvideotemporal}, ReVisionLLM~\cite{hannan2024revisionllmrecursivevisionlanguagemodel}, ReasonAct~\cite{liu2025reasonactprogressivetrainingfinegrained}, ST-Think~\cite{wu2025stthinkmultimodallargelanguage}, Seed1.5-VL~\cite{guo2025seed15vltechnicalreport}; The complete list can be found in \Cref{tab:video_llms_normal_merged}.}, answer_work]
            ]
        ]
        [Reinforcement Learning (RL)~\S~\ref{sec:rl}, for tree={fill=red!45, template}
            [RL for Alignment, template
                [{VistaDPO~\cite{huang2025vistadpo}, Video-MTR \cite{xie2025videomtrreinforcedmultiturnreasoning}, video-SALMONN-o1 \cite{sun2025videosalmonno1reasoningenhancedaudiovisuallarge}, LLaVA-NeXT-Video-Thinking \cite{gao2025exploringhallucinationlargemultimodal}, VerIPO (Verifier-DPO) \cite{li2025veripocultivatinglongreasoning}}, template_work]
            ]
            [Video-specific Policy Optimization, template
                [{T-GRPO \cite{feng2025videor1reinforcingvideoreasoning}, Reg-GRPO \cite{park2025deepvideor1videoreinforcementfinetuning}, TW-GRPO \cite{dang2025reinforcingvideoreasoningfocused}, DGRPO \cite{zhang2025thinkingvideosmultimodaltoolaugmented}, Multi-task GRPO \cite{li2025videochatr1enhancingspatiotemporalperception}, vsGRPO \cite{liao2025improvedvisualspatialreasoningr1zerolike}; More details can be found in \Cref{tab:grpo_family_3col_final}.}, template_work]
            ]
            [Reward Design, template
                [{Format/faithfulness \cite{li2025veripocultivatinglongreasoning}, Answer correctness \cite{wang2025videorftincentivizingvideoreasoning}, Temporal localization (tIoU/Recall{@}K) \cite{xu2025viarladaptivetemporalgrounding,luo2025musegreinforcingvideotemporal}, Spatio-temporal grounding (IoU/track/relations) \cite{li2025videochatr1enhancingspatiotemporalperception,ouyang2025spacerreinforcingmllmsvideo}, Budget awareness (viewing/CoT) \cite{chen2025scalingrllongvideos,wang2025videortsrethinkingreinforcementlearning}, Caption dual verifiable rewards \cite{meng2025videocapr1enhancingmllmsvideo}, Audio-aware consistency \cite{sun2025videosalmonno1reasoningenhancedaudiovisuallarge}}, template_work]
            ]
            [RL Data for Video Reasoning, template
                [{Temporal-RLT-490k/32k \cite{li2025reinforcementlearningtuningvideollms}, MTVR-RL-110k \cite{zhang2025thinkingvideosmultimodaltoolaugmented}, Video-R1-260k \cite{feng2025videor1reinforcingvideoreasoning}, LongVideo-Reason-RL~\cite{chen2025scalingrllongvideos}, TVG-RL-18K~\cite{chen2025datasets}}, template_work]
            ]
            [Domain-Specific RL, template
                [{Fact-R1 \cite{zhang2025factr1explainablevideomisinformation}, VAU-R1 \cite{zhu2025vaur1advancingvideoanomaly}, UniVG-R1 \cite{bai2025univgr1reasoningguideduniversal}, VRAgent-R1 \cite{chen2025vragentr1boostingvideorecommendation}, VideoCap-R1 \cite{meng2025videocapr1enhancingmllmsvideo}, VLN-R1 \cite{qi2025vlnr1visionlanguagenavigationreinforcement}, TVG-R1~\cite{chen2025datasets}, AV-Reasoner~\cite{lu2025avreasonerimprovingbenchmarkingcluegrounded}, DAVID-XR1~\cite{gao2025davidxr1detectingaigeneratedvideos}}, template_work]
            ]
            [Reinforced Video-LMMs, template
                [{Video-R1 \cite{feng2025videor1reinforcingvideoreasoning}, VideoChat-R1.5 \cite{yan2025videochat15}, Video-TT \cite{zhang2025videott}, VideoCap-R1~\cite{meng2025videocapr1enhancingmllmsvideo}, Video-RTS \cite{wang2025videortsrethinkingreinforcementlearning}, TinyLLaVA-Video-R1~\cite{chen2025scalingrllongvideos}, SpaceR \cite{ouyang2025spacerreinforcingmllmsvideo}, TimeZero \cite{wang2025timer1posttraininglargevision},  TCoT~\cite{arnab2025temporalchainthoughtlongvideo}, VersaVid-R1~\cite{chen2025versavidr1versatilevideounderstanding}, ViQAgent~\cite{montes2025viqagentzeroshotvideoquestion}, VideoAgent2~\cite{zhi2025videoagent2enhancingllmbasedagent},  VideoForest~\cite{meng2025videoforestpersonanchoredhierarchicalreasoning}, VideoMind~\cite{liu2025videomindchainofloraagentlong}
; The complete list can be found in \Cref{tab:video_llms_normal_merged}.}, template_work]
            ]
        ]
        [Test-Time Scaling (TTS)~\S~\ref{sec:tts}, for tree={fill=blue!35, multiple}
            [Video Chain-of-Thought Prompting, multiple
                [{Video-of-Thought \cite{fei2024videoofthoughtstepbystepvideoreasoning}, CoT-Vid \cite{jin2025cotviddynamicchainofthoughtrouting}, Video-Skill-CoT~\cite{lee2025videoskillcotskillbasedchainofthoughtsdomainadaptive}, AKeyS~\cite{fan2025agentickeyframesearchvideo}, CoS~\cite{hu2025coschainofshotpromptinglong}, NoWait~\cite{wang2025waitdontneedwait}, ReVisionLLM~\cite{hannan2024revisionllmrecursivevisionlanguagemodel}, TCoT~\cite{arnab2025temporalchainthoughtlongvideo}, VCA~\cite{yang2025vcavideocuriousagent}, VidChain~\cite{lee2025vidchainchainoftasksmetricbaseddirect}, VideoDeepResearch~\cite{yuan2025videodeepresearchlongvideounderstanding}}, multiple_work]
            ]
            [Self-Consistency Decoding, multiple
                [{Multi-path sampling + voting (CoT-Vid) \cite{jin2025cotviddynamicchainofthoughtrouting}}, multiple_work]
            ]
            [Confidence-Based Iterative Reasoning, multiple
                [{CyberV \cite{meng2025cybervcyberneticstesttimescaling}, Video-ICL \cite{kim2024videoiclconfidencebasediterativeincontext}}, multiple_work]
            ]
            [Self-Improvement via Refinement Loops, multiple
                [{DIVE \cite{kamoto2025divedeepsearchiterativevideo}, Video-MTR \cite{xie2025videomtrreinforcedmultiturnreasoning}}, multiple_work]
            ]
            [MCTS for Video Captioning, multiple
                [{AutoCaption / MCTS-VCB \cite{yu2025evaluatingmllmsvideocaptioningmcts}}, multiple_work]
            ]
            [Multi-Path \& Routing, multiple
                [{MR.\,Video \cite{pang2025mrvideomapreduceprinciple}, Free-MoRef \cite{wang2025freemorefinstantlymultiplexingcontext}, SiLVR \cite{zhang2025silvrsimplelanguagebasedvideo}}, multiple_work]
            ]
            [Tool-Augmented Reasoning, multiple
                [{VITAL \cite{zhang2025thinkingvideosmultimodaltoolaugmented}, Pixel-Reasoner~\cite{su2025pixelreasonerincentivizingpixelspace}, Ego-R1 \cite{tian2025egor1chainoftoolthoughtultralongegocentric}, ReAgent-V~\cite{zhou2025reagentvrewarddrivenmultiagentframework}, VideoDeepResearch \cite{yuan2025videodeepresearchlongvideounderstanding}, Agentic Keyframe Search \cite{fan2025agentickeyframesearchvideo}, 
                VideoExplorer~\cite{yuan2025videodeepresearch}}, multiple_work]
            ]
        ]
        [Benchmarks for Video Reasoning~\S~\ref{sec:benchmarks}, for tree={fill=purple!70, tuning}
            [General Video QA Benchmarks, tuning
                [{MMVU \cite{zhao2025mmvumeasuringexpertlevelmultidiscipline}, MVBench~\cite{li2024mvbench}, NeXT-QA~\cite{xiao2021nextqa}, VideoMME~\cite{fu2025videomme}, VidTAB-QA~\cite{li2024videoeval}, Dream-1k~\cite{wang2024dream1k}, VidComposition~\cite{tang2024vidcompostion}}, tuning_work]
            ]
            [Video Reasoning Benchmarks, tuning
                [{VCR-Bench \cite{qi2025vcrbenchcomprehensiveevaluationframework}, VideoReasonBench \cite{liu2025videoreasonbenchmllmsperformvisioncentric}, MINERVA \cite{nagrani2025minervaevaluatingcomplexvideo}, MECD \cite{chen2024mecdunlockingmultieventcausal}, HAVEN \cite{gao2025exploringhallucinationlargemultimodal}, VidHalluc \cite{li2025vidhallucevaluatingtemporalhallucinations}}, tuning_work]
            ]
            [Grounding Reasoning Benchmarks, tuning
                [{Charades-STA, ActivityNet-Grounding, ActivityNet-RTL \cite{li2025reinforcementlearningtuningvideollms}, V-STAR \cite{cheng2025vstarbenchmarkingvideollmsvideo}, VSI-Bench \cite{ouyang2025spacerreinforcingmllmsvideo}, NExT-GQA~\cite{xiao2024nextgqa}, GoT--10k~\cite{huang2019got}}, tuning_work]
            ]
            [Long/Streaming Evaluation, tuning
                [{LongVideo-Reason-eval \cite{chen2025scalingrllongvideos}, HLV-1K \cite{zou2025hlv1klargescalehourlongvideo}, ScaleLong \cite{ma2025scalelongmultitimescalebenchmarklong}, SVBench \cite{yang2025svbenchbenchmarktemporalmultiturn}, CogStream \cite{zhao2025cogstreamcontextguidedstreamingvideo}, OmniMMI \cite{wang2025omnimmicomprehensivemultimodalinteraction}}, tuning_work]
            ]
        ]
    ]
\end{forest}
\caption{Taxonomy of Video-LMM post-training.}
\label{fig:taxonomy_videoposttraining}
\end{figure}
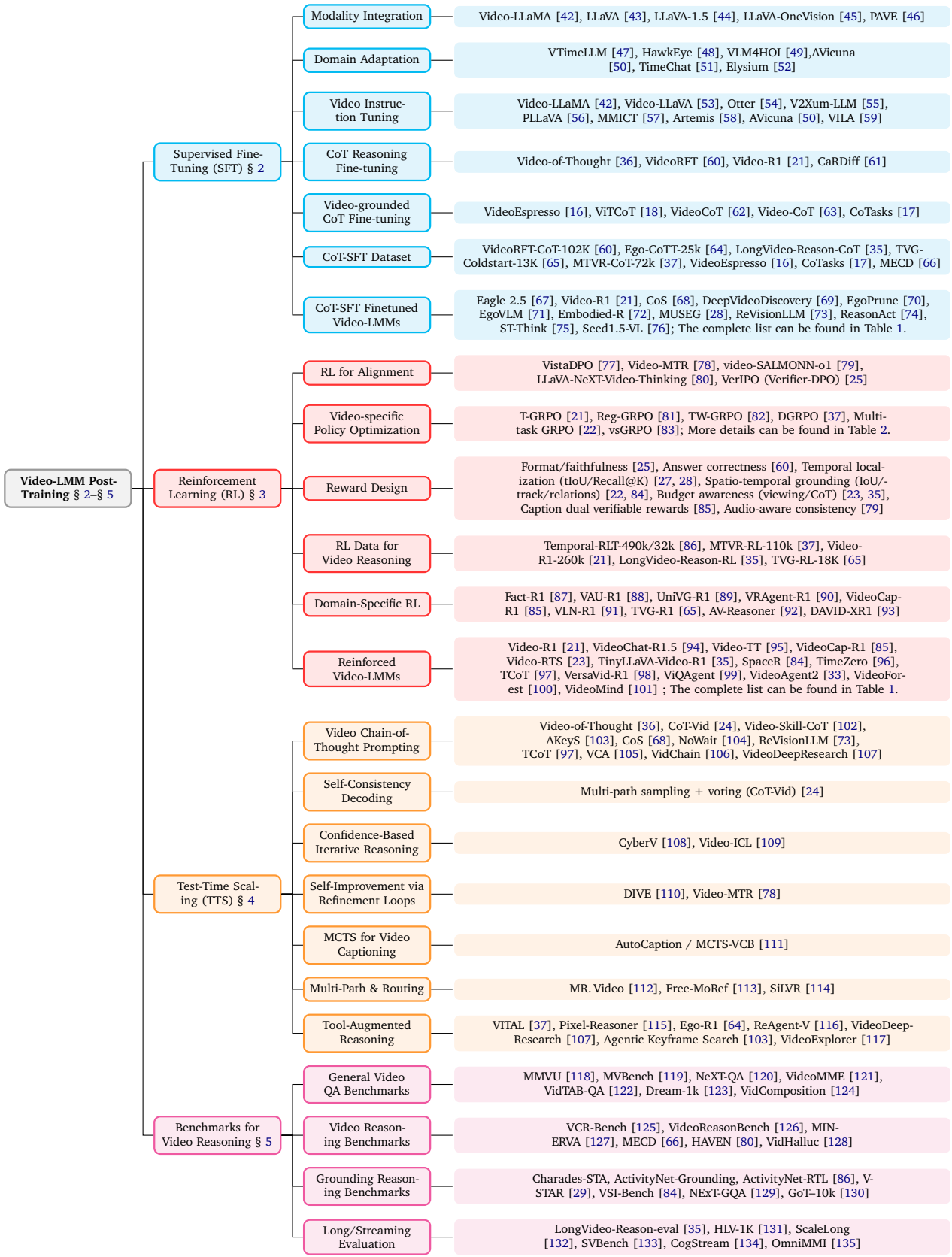

\subsection{Basic SFT for Video-LMMs}

Researchers have discovered large-scale pretraining methods that enable LLMs to effectively consume internet-scale unlabeled text corpora through next token prediction, trained with maximum likelihood estimation (MLE) to obtain powerful LLM base models. These base models are then further refined through SFT using high-quality annotated data in smaller quantities. Early SFT for text-only LLMs primarily served two purposes: enhancing the model's instruction-following capability and performing domain adaptation to transform general-purpose LLMs into domain-specific experts. For obtaining an LMM, subsequent SFT can either build upon the LLM base model or start from an instruction-tuned LLM for further refinement.

\paragraph{Modality Integration.} The transition from LLM to LMM typically begins with a Modality Integration stage, which endows the LLM with the ability to understand information from other modalities, particularly visual information. This stage usually employs large-scale image-text pairs for image captioning tasks, sometimes incorporating video-text pairs as well. A connector links the vision encoder to the LLM, and supervised fine-tuning is applied to update either the connector parameters alone or both the connector and LLM parameters jointly. The connector is typically a linear layer or MLP that maintains input-output token correspondence, though alternatives like Q-Former~\cite{li2023blip} use resamplers to map inputs to a fixed number of tokens. In practice, the former approach generally outperforms the latter~\cite{llava}. Additionally, some methods directly feed vision features to the LLM, potentially passing representations from different ViT layers to corresponding LLM layers. Regardless of the specific approach, the key objective of modality integration is to effectively project visual representations from the vision encoder into the LLM's embedding space, enabling the LMM to directly interpret visual information. Beyond vision, other modalities such as audio, speech, and optical flow can be aligned with LMMs using similar operations~\cite{videollama}.

\paragraph{Domain Adaptation.} Domain adaptation in Video-LMMs can be understood in multiple ways. The most fundamental interpretation applies when an LMM has only performed modality integration on image-text data without extending to video: an additional domain adaptation step uses video-text pairs to fine-tune the LMM for video captioning, thereby expanding the LMM's capabilities to video understanding. A second interpretation involves a Video-LMM that initially handles only general video understanding being fine-tuned with domain-specific data to inject domain knowledge, enabling it to process specialized content such as medical videos, anomaly detection videos, or AI-generated video detection. A third interpretation involves endowing Video-LMMs with specific capabilities, such as temporal localization abilities. For instance, VTimeLLM~\cite{huang2024vtimellm}, TimeChat~\cite{ren2023timechat}, and AVicuna~\cite{tang2024avicuna} employ boundary alignment to align events occurring in videos with their start and end times, enabling LMMs to predict when events occur in videos. Elysium~\cite{wang2024elysium} extends this capability to the spatiotemporal domain. Research indicates that domain adaptation may compromise the instruction-following ability inherited from the LLM, typically necessitating further SFT to restore this capability.

\paragraph{Video Instruction Tuning.} Video Instruction tuning enhances the instruction-following capability of Video-LMMs~\cite{song2023moviechat,maaz2023video,chen2023videochatcap, chen2023videollm,Ma2023Vista-LLaMA,Zhang2024LLoVi,xu2024pllava,munasinghe2023pg}. The training data takes the form of instruction-response pairs, and after fine-tuning, the model is expected to respond as accurately as possible to any given instruction~\cite{kumar2025llm}. For example, when asked to provide a video-to-text summarization of a video, the model generates a description; when asked for video-to-video summarization, the model outputs the indices of key frames~\cite{hua2024v2xum}. Visual instruction tuning originated with LLaVA~\cite{llava} and typically follows the Modality Integration stage, though some work has shown that mixing modality integration data with instruction tuning data in a unified format yields better results~\cite{llava15}. Video-LLaMA extended instruction tuning to video and audio, validating the feasibility of video instruction tuning~\cite{videollama}. Since then, instruction tuning has been widely applied to video understanding~\cite{weng2024longvlm,Wang2023VaQuitA,zhang2024longva,cheng2024videollama,zhang2025videollama3,jin2024chatuniviunifiedvisualrepresentation,xu2024slowfastllava,zhou2025streferempoweringvideollms,wang2025timetemporalsensitivemultidimensionalinstruction,liao2024videoinstazeroshotlongvideo,huang2024prunevidvisualtokenpruning,jiang2025stormtokenefficientlongvideo,cheng2025tempuratemporaleventmasked,sun2024video,maaz2024videogpt+}.

These fine-tuning approaches all employ auto-regressive language modeling loss as the objective function. While full fine-tuning of the LLM is possible, it can be computationally and memory-intensive, leading to frequent adoption of parameter-efficient fine-tuning (PEFT) techniques. For example, some approaches only update LoRA~\cite{hu2022lora} and connector parameters, while others attempt to fine-tune the vision encoder. Input prompts typically include video placeholders that are replaced with corresponding video tokens before being fed into the LLM.

\subsection{From Video Instruction Tuning to Chain-of-Thought Fine-tuning (CoT-SFT)}

\paragraph{CoT Reasoning Fine-tuning.}
Chain-of-thought (CoT) reasoning emphasizes introducing additional intermediate steps to improve final answer accuracy, requiring models to output step-by-step reasoning traces. Research has found that longer CoTs not only provide interpretability but also enhance final answer accuracy, a phenomenon that will be further discussed in \Cref{sec:tts}. CoT reasoning fine-tuning uses data in long CoT format (either annotated by human experts or generated synthetically) and applies the same supervised training methodology as instruction tuning to internalize the capability of producing step-by-step reasoning traces into the model. This approach can also be extended to the multimodal domain. For example, the CoT data in Video-of-Thought~\cite{fei2024videoofthoughtstepbystepvideoreasoning} divides the process of answering a video QA question into five steps: analyzing the user's question, constructing a scene graph of the input video, generating detailed video captions, using the acquired information to analyze which option is optimal by comparing against the question and choices, and finally summarizing the entire reasoning process to return the answer.

\paragraph{Video-grounded CoT Fine-tuning.} Early CoT reasoning fine-tuning took video and prompts as input and produced pure text as output, which to some extent limited the capabilities of Video-LMMs. Text-only CoTs emphasize logical structure but risk visual hallucination. Therefore, incorporating vision-grounded information into CoTs is beneficial. Video-grounded CoTs reduce hallucination by binding steps to visual evidence via timestamps, shot IDs, or frame indices. VideoEspresso~\cite{han2024videoespressolargescalechainofthoughtdataset} demonstrates that pairing CoT with core frame selection yields fine-grained reasoning supervision while controlling token budgets. ViTCoT~\cite{zhang2025vitcotvideotextinterleavedchainofthought} advocates video-text interleaving during reasoning, periodically revisiting key frames while thinking, to better align cognition with perception. CoTasks~\cite{wang2025cotaskschainofthoughtbasedvideo} further structures the reasoning interface by injecting entity-level intermediate steps (localization, tracking, relation extraction) as part of the supervision, improving compositional spatiotemporal reasoning.

\paragraph{CoT Fine-tuning for Video RL Cold-Start.}

Although works such as Video-of-Thought~\cite{fei2024videoofthoughtstepbystepvideoreasoning} and VideoEspresso~\cite{han2024videoespressolargescalechainofthoughtdataset} have achieved certain success in introducing CoT to video reasoning, the CoT formats used in these datasets are typically fixed, following rigid step sequences. While unified formats facilitate batch generation, they consequently lack flexibility: models cannot explore independently, and predefined paths may not be optimal. Errors generated during fixed-path reasoning cannot be effectively corrected and accumulate continuously. Fundamentally, this represents a static learning paradigm whose effectiveness is highly dependent on the quality and diversity of training data. These models can only imitate the reasoning patterns present in their dataset and struggle to generalize to unseen, more complex scenarios~\cite{zhou2025reasoningneedvideogeneralization}. To overcome this limitation and enable models to learn and align with more abstract and qualitative objectives that are difficult to define precisely in a supervised dataset, many works are increasingly turning to Reinforcement Learning (RL, which will be detailed in~\label{sec:rl}), particularly following the emergence of R1-style and GRPO algorithms. Consequently, CoT-SFT has gradually evolved into the cold-start training phase for RL. The cold-start phase is now critical for stabilizing the model before full RL training, preventing instability that can arise from purely RL-driven updates. Cold-start data preparation focuses on capturing human-readable reasoning patterns to prevent instability from purely RL-driven updates. This step generates CoT-style examples with consistent <think> and <answer> fields, usually involving thousands of carefully curated samples. Structured CoT formats and consistent fields ensure clarity and robustness in the model's reasoning outputs, reducing errors and improving interpretability~\cite{kumar2025llm}.

\subsection{Data Construction and Representative Resources}

\paragraph{Curation Pipelines.} Obtaining high-quality video CoT supervision is resource-intensive. A practical approach involves a two-phase curation pipeline: (1) eliciting preliminary CoTs from a reasoning-capable LLM using structured video metadata such as scene descriptions, automatic speech recognition (ASR) transcripts, and shot lists; (2) applying visual consistency refinement through an LMM conditioned on actual video frames to reduce hallucination and align reasoning steps with visual evidence. VideoRFT~\cite{wang2025videorftincentivizingvideoreasoning} exemplifies this methodology and provides the VideoRFT-CoT-102K dataset for SFT alongside larger collections designed for RL training.

\paragraph{CoT-SFT Datasets for Video Reasoning.}
We highlight representative resources used for SFT with CoT format. VideoRFT-CoT-102K supplies large-scale CoT traces tailored for reward-driven fine-tuning and incentivized video reasoning~\cite{wang2025videorftincentivizingvideoreasoning}. PixelReasoner-SFT offers pixel/region-grounded, stepwise supervision that tightly couples perception with structured reasoning. Ego-CoTT-25k targets egocentric and embodied scenarios with chain-of-tool-thought style supervision for ultra-long videos~\cite{tian2025egor1chainoftoolthoughtultralongegocentric}. LongVideo-Reason-CoT~\cite{chen2025scalingrllongvideos} extends to multi-event, long-form understanding with narrative-level annotations and supports long-context training pipelines. MTVR-CoT-72k~\cite{zhang2025thinkingvideosmultimodaltoolaugmented}, including MTVR-CoT and MTVR-CoT-Tool, contribute multi-task CoT trajectories that bridge video QA and temporal grounding, enabling explicit intermediate reasoning. Beyond the above, fine-grained CoT resources such as VideoEspresso~\cite{han2024videoespressolargescalechainofthoughtdataset}, entity-centric CoTasks~\cite{wang2025cotaskschainofthoughtbasedvideo}, and interleaved video–text protocols ViTCoT/ViTIB~\cite{zhang2025vitcotvideotextinterleavedchainofthought} are widely used as warm-up data, while causal/multi-event understanding can leverage MECD~\cite{chen2024mecdunlockingmultieventcausal}. In addition, Video-of-Thought style collections and their perception-to-cognition protocols provide useful templates for supervising intermediate steps~\cite{fei2024videoofthoughtstepbystepvideoreasoning}. More resources are summarized in \Cref{tab:data_summary_pt}.

\paragraph{Long-Video Considerations.} For long-form video content, SFT typically combines CoT supervision with token-budget control mechanisms, such as shot selection and quota assignment, to maintain computational tractability. These approaches may leverage agentic keyframe selection strategies or frame-aware reasoning signals~\cite{ghazanfari2025chainofframesadvancingvideounderstanding}. When SFT precedes RL training on long videos, as demonstrated in LongVILA-R1~\cite{chen2025scalingrllongvideos}, CoT-SFT establishes the format prior that facilitates efficient rollouts and subsequent policy optimization~\cite{chen2025scalingrllongvideos}.

\begin{table}[!ht]
\centering
\caption{Summary of large multimodal models for video reasoning (Video-LMMs), including model name, number of parameters, training strategy, test-time scaling, and links.}
\label{tab:video_llms_normal_merged}
\resizebox{0.9\linewidth}{!}{%
\begin{tabular}{>{\bfseries}p{4.6cm} p{1.8cm} p{1.5cm} p{8cm} p{0.5cm} p{1.5cm}}
\toprule
\textbf{Model} & \textbf{\# Params} & \textbf{\# Stages} & \textbf{Training Strategy} & \textbf{TTS} & \textbf{Link} \\
\midrule
Fact-R1~\cite{zhang2025factr1explainablevideomisinformation} & $\sim$7B &3 & SFT + DPO + GRPO & \cmark & \href{https://github.com/zfr00/fact-r1}{\faGithub}\; \href{https://huggingface.co/papers/2505.16836}{\faHuggingFace}\\
\midrule
Temporal-RLT~\cite{li2025reinforcementlearningtuningvideollms} & 7B &2 & SFT + GRPO & \cmark & 
\href{https://github.com/appletea233/Temporal-R1}{\faGithub}\; \href{https://huggingface.co/papers/2506.01908}{\faHuggingFace}
\\
\midrule
VideoChat-R1~\cite{li2025videochatr1enhancingspatiotemporalperception} & 7B & 1 & Multi-task RFT (GRPO) & \cmark & 
\href{https://github.com/OpenGVLab/VideoChat-R1}{\faGithub}\; \href{https://huggingface.co/OpenGVLab/VideoChat-R1_5-7B}{\faHuggingFace}
\\
\midrule
Spatial-R1~\cite{ouyang2025spacerreinforcingmllmsvideo} & 7B & 1 & Task-Specific RFT (GRPO) & \xmark & 
\href{https://github.com/OuyangKun10/SpaceR}{\faGithub}\; \href{https://huggingface.co/RUBBISHLIKE/SpaceR}{\faHuggingFace}
\\
\midrule
LLaVA-NeXT-Video-Thinking~\cite{gao2025exploringhallucinationlargemultimodal} & 7B - 34B &2 & SFT + TDPO (RLHF-style, Segment-Weighted) & \cmark & 
\href{https://github.com/Hongcheng-Gao/HAVEN}{\faGithub}\; \href{https://huggingface.co/datasets/Joshua999/HAVEN}{\faHuggingFace}
\\
\midrule
video-SALMONN-o1~\cite{sun2025videosalmonno1reasoningenhancedaudiovisuallarge} & 7B &2 & SFT (LoRA) + pDPO (Process-level) & \cmark & 
\href{https://github.com/BriansIDP/video-SALMONN-o1}{\faGithub}\; \href{https://huggingface.co/tsinghua-ee/video-SALMONN-o1}{\faHuggingFace}
\\
\midrule
LongVILA-R1~\cite{chen2025scalingrllongvideos} & 7B - 8B &2 & CoT-SFT + RL (MR-SP, GRPO) & \xmark & 
\href{https://github.com/NVlabs/Long-RL}{\faGithub}\; \href{https://huggingface.co/Efficient-Large-Model/LongVILA-R1-7B}{\faHuggingFace}
\\
\midrule
Video-RTS~\cite{wang2025videortsrethinkingreinforcementlearning} & 7B & 1 & Pure RL, no SFT (GRPO) & \cmark & 
\href{https://github.com/Ziyang412/Video-RTS}{\faGithub}\; \href{https://huggingface.co/Ted412/Video-RTS}{\faHuggingFace}
\\
\midrule
Ego-R1~\cite{tian2025egor1chainoftoolthoughtultralongegocentric} & $\sim$3B &2 & SFT (CoTT) + RL (GRPO) & \cmark & 
\href{https://github.com/egolife-ai/Ego-R1}{\faGithub}\; \href{https://huggingface.co/Ego-R1}{\faHuggingFace}
\\
\midrule
DeepVideo-R1~\cite{park2025deepvideor1videoreinforcementfinetuning} & 2B - 7B & 1 & Regressive GRPO (Reg-GRPO) & \cmark & 
\href{https://github.com/mlvlab/DeepVideoR1}{\faGithub}\; \href{https://huggingface.co/papers/2506.07464}{\faHuggingFace}
\\
\midrule
VideoRFT~\cite{wang2025videorftincentivizingvideoreasoning} & $\sim$7B &2 & SFT (CoT) + RL (GRPO) & \cmark &
\href{https://github.com/QiWang98/VideoRFT}{\faGithub}\; \href{https://huggingface.co/QiWang98/VideoRFT}{\faHuggingFace}
\\
\midrule
UniVG-R1~\cite{bai2025univgr1reasoningguideduniversal} & 2B - 7B &2 & CoT-SFT + RL (GRPO) & \cmark & 
\href{https://github.com/AMAP-ML/UniVG-R1}{\faGithub}\; \href{https://huggingface.co/GD-ML/UniVG-R1}{\faHuggingFace}
\\
\midrule
TinyLLaVA-Video-R1~\cite{zhang2025tinyllavavideor1smallerlmmsvideo} & $\sim$3B &2 & SFT (Cold-Start) + RL (GRPO) & \cmark & 
\href{https://github.com/ZhangXJ199/TinyLLaVA-Video-R1}{\faGithub}\; \href{https://huggingface.co/Zhang199/TinyLLaVA-Video-R1}{\faHuggingFace}
\\
\midrule
Video-R1~\cite{feng2025videor1reinforcingvideoreasoning} & 7B &2 & CoT-SFT + RL (Temporal GRPO) & \cmark & 
\href{https://github.com/tulerfeng/Video-R1}{\faGithub}\; \href{https://huggingface.co/Video-R1/Video-R1-7B}{\faHuggingFace}
\\
\midrule
VAU-R1~\cite{zhu2025vaur1advancingvideoanomaly} & 2B - 3B &2 & SFT + RFT (GRPO) & \cmark & 
\href{https://github.com/GVCLab/VAU-R1}{\faGithub}\; \href{https://huggingface.co/datasets/7xiang/VAU-Bench}{\faHuggingFace}
\\
\midrule
ST-R1~\cite{wu2025stthinkmultimodallargelanguage} & $\sim$7B &2 & CoT-SFT + RL (GRPO) & \cmark & 
\href{https://github.com/WPR001/Ego-ST}{\faGithub}\; \href{https://huggingface.co/collections/openinterx/st-think-684745a22833bfda73c5bd82}{\faHuggingFace}
\\
\midrule
TimeZero~\cite{wang2025timer1posttraininglargevision} & $\sim$7B & 1 & Pure RL (GRPO) & \cmark & 
\href{https://github.com/www-Ye/Time-R1}{\faGithub}\; \href{https://huggingface.co/wwwyyy/TimeZero-Charades-7B}{\faHuggingFace}
\\
\midrule
VerIPO~\cite{li2025veripocultivatinglongreasoning} & 7B &3 & GRPO-Verifier-DPO loop & \cmark & 
\href{https://github.com/HITsz-TMG/VerIPO}{\faGithub}\; \href{https://huggingface.co/Uni-MoE/VerIPO-7B-v1.0}{\faHuggingFace}
\\
\midrule
VLN-R1~\cite{qi2025vlnr1visionlanguagenavigationreinforcement} & $\sim$7B &2 & SFT + RFT (Custom Reward) & \xmark & 
\href{https://github.com/Qi-Zhangyang/GPT4Scene-and-VLN-R1}{\faGithub}\; \href{https://huggingface.co/datasets/alexzyqi/VLN-Ego}{\faHuggingFace}
\\
\midrule
TVG-R1~\cite{chen2025datasets} & $\sim$7B &2 & SFT + RFT & \cmark & 
\href{https://github.com/zjuruizhechen/TVG-R1}{\faGithub}\; \href{https://huggingface.co/RuizheChen/TVG-R1}{\faHuggingFace}
\\
\midrule
VideoCap-R1~\cite{meng2025videocapr1enhancingmllmsvideo} & $\sim$7B &2 & SFT + RL (GRPO) & \cmark & --\\
\midrule
Vad-R1~\cite{huang2025vadr1videoanomalyreasoning} & $\sim$7B &2 & P2C-CoT SFT + AVA-GRPO & \cmark & 
\href{https://github.com/wbfwonderful/Vad-R1}{\faGithub}\; \href{https://huggingface.co/datasets/wbfwonderful/Vad-R1}{\faHuggingFace}
\\
\midrule
R1-SGG~\cite{chen2025compilescenegraphsreinforcement} & 2B - 7B &2 & SFT + RL (GRPO) & \xmark & 
\href{https://github.com/gpt4vision/R1-SGG}{\faGithub}\; \href{https://huggingface.co/JosephZ/R1-SGG-7B}{\faHuggingFace}
\\
\midrule
vsGRPO~\cite{liao2025improvedvisualspatialreasoningr1zerolike} & 2B - 7B & 1 & R1-Zero-like RL training (GRPO) & \cmark & 
\href{https://github.com/zhijie-group/R1-Zero-VSI}{\faGithub}\; \href{https://huggingface.co/datasets/OPPOer/VSI-100k}{\faHuggingFace}
\\
\midrule
BusterX~\cite{wen2025busterxmllmpoweredaigeneratedvideo} & $\sim$7B &2 & SFT (Cold-start) + RL (PEFT, DAPO) & \cmark & 
\href{https://github.com/l8cv/BusterX}{\faGithub}\; \href{https://huggingface.co/l8cv/BusterX_plusplus}{\faHuggingFace}
\\
\midrule
ARC-Hunyuan-Video~\cite{ge2025archunyuanvideo7bstructuredvideocomprehension} & 7B &4 & SFT + CoT SFT + RL (GRPO) + SFT & \xmark & 
\href{https://github.com/TencentARC/ARC-Hunyuan-Video-7B/}{\faGithub}\; \href{https://huggingface.co/TencentARC/ARC-Hunyuan-Video-7B}{\faHuggingFace}
\\
\midrule
VITAL~\cite{zhang2025thinkingvideosmultimodaltoolaugmented} & 7B &7 & SFT + Tool-Augmented DGRPO & \cmark & 
\href{https://github.com/zhang9302002/ThinkingWithVideos}{\faGithub}\; \href{https://huggingface.co/datasets/zhang9302002/MultiTaskVideoReasoning}{\faHuggingFace}
\\
\midrule
Video-MTR~\cite{xie2025videomtrreinforcedmultiturnreasoning} & $\sim$7B & 1 & RL with Gated Bi-Level Reward (PPO) & \xmark & 
- \href{https://huggingface.co/papers/2508.20478}{\faHuggingFace}
\\
\midrule
ReasonAct~\cite{liu2025reasonactprogressivetrainingfinegrained} & 3B &3 & SFT + V-SFT + Temporal RL (T-GRPO) & \xmark & --\\
\midrule
ReFoCUS~\cite{lee2025refocusreinforcementguidedframeoptimization} & - &2 & RL with Reward Model (GRPO) & \xmark & 
- \href{https://huggingface.co/papers/2506.01274}{\faHuggingFace}
\\
\midrule
Kwai Keye-VL~\cite{kwaikeyeteam2025kwaikeyevltechnicalreport} & 8.4B &2 & SFT + MPO + Mix-Mode RL (MPO, GRPO) & \cmark & 
\href{https://github.com/Kwai-Keye/Keye}{\faGithub}\; \href{https://huggingface.co/Kwai-Keye/Keye-VL-8B-Preview}{\faHuggingFace}
\\
\midrule
VRAgent-R1~\cite{chen2025vragentr1boostingvideorecommendation} & - &2 & Progressive RL for User Simulation (GRPO) & \cmark & --\\
\midrule
Omni-R1~\cite{zhong2025omnir1reinforcementlearningomnimodal} & 7B &2 & End-to-End RL (GRPO) & \xmark & 
\href{https://github.com/aim-uofa/Omni-R1}{\faGithub}\; \href{https://huggingface.co/Haoz0206/Omni-R1}{\faHuggingFace}
\\
\midrule
A2Seek-R1~\cite{mo2025a2seekreasoningcentricbenchmarkaerial} & $\sim$3B &2 & GoT-SFT + RFT (Aerial GRPO) & \xmark & 
\href{https://hayneyday.github.io/A2Seek/}{\faGithub}\;-
\\
\midrule
Pixel Reasoner~\cite{su2025pixelreasonerincentivizingpixelspace} & 7B &2 & SFT + Curiosity-Driven RL (Custom) & \cmark & 
\href{https://github.com/TIGER-AI-Lab/Pixel-Reasoner}{\faGithub}\; \href{https://huggingface.co/TIGER-Lab/PixelReasoner-RL-v1}{\faHuggingFace}
\\
\midrule
Tempo-R0~\cite{yue2025tempor0videomllmtemporalvideo} & $\sim$7B &2 & SFT + RFT (PIR-GRPO) & \xmark & --\\
\midrule
VideoSafety-R1~\cite{sun2025evaluationdefenseadvancingsafety} & - & 2 & AT-SFT + RLHF-style (GRPO) & \cmark & --\\
\midrule
SiLVR~\cite{zhang2025silvrsimplelanguagebasedvideo} & 7B - 72B & N/A & Training-Free, Modular & \xmark & 
\href{https://github.com/CeeZh/SILVR}{\faGithub}\;-
\\
\midrule
CoT-Vid~\cite{jin2025cotviddynamicchainofthoughtrouting} & 7B & N/A & Training-Free, Inference-time strategy & \cmark & --\\
\midrule
MR. Video~\cite{pang2025mrvideomapreduceprinciple} & Modular & N/A & Training-Free, MapReduce Framework & \cmark & 
\href{https://github.com/ziqipang/MR-Video}{\faGithub}\; \href{https://huggingface.co/ziqipang/MR-Video}{\faHuggingFace}
\\
\midrule
Free-MoRef~\cite{wang2025freemorefinstantlymultiplexingcontext} & 7B & N/A & Training-Free, Inference-time MoE & \xmark & 
\href{https://github.com/wkfdb/Free-MoRef}{\faGithub}\;-
\\
\bottomrule
\end{tabular}
}
\end{table}
\section{Reinforcement Learning for Video Reasoning}\label{sec:rl}

\begin{myboxi}[Takeaways]
\begin{itemize}
\item GRPO has emerged as a popular approach in recent work on video reasoning because it uses verifiable outcomes like answer correctness for optimization, avoiding the need for human preference data.
\item A successful system requires co-designing three key elements: advanced policy algorithms, multi-faceted reward functions, and high-quality curated datasets.
\item This reinforcement learning approach is highly data-efficient, as a small set of quality data can match or exceed the performance of large-scale supervised tuning.
\end{itemize}
\end{myboxi}

\subsection{Preliminary: From PPO to GRPO}

This subsection formalizes three alignment routes that underpin post-training for video reasoning: PPO-based RLHF (with or without AI-generated preferences), Direct Preference Optimization (DPO), and Group Relative Policy Optimization (GRPO). We use $x$ for the multimodal context, $y$ for a response, and $\tau$ for a token trajectory.

\paragraph{PPO, RLHF, and RLAIF.}
RLHF trains a reward model (RM) to score responses and then optimizes the policy with PPO under a KL constraint to a reference model. The RM is commonly trained on preference pairs $(x, y^{+}, y^{-})$ via a Bradley–Terry objective,
\[
\mathcal{L}_{\mathrm{RM}}(\phi)
= - \mathbb{E}_{(x,y^{+},y^{-})}\,
\log \sigma\!\left(r_{\phi}(x,y^{+}) - r_{\phi}(x,y^{-})\right),
\]
where $r_{\phi}(x,y)$ is the scalar reward and $\sigma$ is the logistic function. Given a fixed RM, PPO maximizes a clipped policy-gradient objective with a KL penalty to the reference $\pi_{\mathrm{ref}}$ (\eg, SFT model). Let $r_t(\theta)=\frac{\pi_{\theta}(y_t\,|\,x,y_{<t})}{\pi_{\theta_{\mathrm{old}}}(y_t\,|\,x,y_{<t})}$ and $\hat{A}_t$ be an advantage estimator (often sequence-level reward broadcast to tokens):
\[
\mathcal{L}_{\mathrm{PPO}}(\theta)
= -\mathbb{E}\bigg[\sum_{t}
\min\!\Big(r_t(\theta)\,\hat{A}_t,\;
\mathrm{clip}\big(r_t(\theta),\,1-\epsilon,\,1+\epsilon\big)\,\hat{A}_t\Big)\bigg]
\;+\; \beta\,\mathrm{KL}\!\left(\pi_{\theta}(\cdot|x)\,\|\,\pi_{\mathrm{ref}}(\cdot|x)\right).
\]
RLAIF replaces human preferences with AI-generated preferences or rewards; the optimization is unchanged, only the supervision source for $\mathcal{L}_{\mathrm{RM}}$ differs. In our curated corpus of video-LLM papers, explicit post-training with PPO/RLHF/RLAIF is uncommon relative to DPO/GRPO.

\paragraph{Direct Preference Optimization (DPO).}
DPO dispenses with an explicit RM and directly matches the policy to the observed preferences relative to a fixed reference model $\pi_{\mathrm{ref}}$. With temperature $\beta>0$, the standard DPO loss over $(x,y^{+},y^{-})$ is
\[
\mathcal{L}_{\mathrm{DPO}}(\theta)
= -\,\mathbb{E}\,\log \sigma\!\Big(
\beta\big[
\log \pi_{\theta}(y^{+}\!|x) - \log \pi_{\mathrm{ref}}(y^{+}\!|x)
- \log \pi_{\theta}(y^{-}\!|x) + \log \pi_{\mathrm{ref}}(y^{-}\!|x)
\big]\Big).
\]
Equivalently, DPO can be viewed as maximizing the log-odds that the policy assigns higher normalized preference to $y^{+}$ than to $y^{-}$, implicitly inducing a reward proportional to $\log \pi_{\theta}(\cdot|x)-\log \pi_{\mathrm{ref}}(\cdot|x)$. Recent video-LLMs instantiate this route with process-/task-aware variants, including video-SALMONN-o1 (process-DPO)~\cite{sun2025videosalmonno1reasoningenhancedaudiovisuallarge}, Fact-R1 (preference stage)~\cite{zhang2025factr1explainablevideomisinformation}, and LLaVA-NeXT-Video-7B-Thinking (TDPO)~\cite{gao2025exploringhallucinationlargemultimodal}.

\paragraph{Group Relative Policy Optimization (GRPO).}
GRPO replaces learned rewards with verifiable outcome rules and optimizes with group-relative advantages. For each prompt $x$, sample $K$ trajectories $\{\tau^{(k)}\}_{k=1}^{K}$ from $\pi_{\theta_{\mathrm{old}}}$, compute verifiable scores $r^{(k)}\in[0,1]$ (\eg, answer correctness, temporal IoU, format checks), and form the group baseline $\bar{r}=\frac{1}{K}\sum_{j=1}^{K}r^{(j)}$. Define advantages
\[
A^{(k)} = r^{(k)} - \operatorname{stopgrad}(\bar{r}),
\qquad
\ell^{(k)}(\theta)=\sum_{t\in \tau^{(k)}} \log \pi_{\theta}(y_t\,|\,x,y_{<t}),
\]
and optimize a KL-regularized objective,
\[
\mathcal{L}_{\mathrm{GRPO}}(\theta)
= -\frac{1}{K}\sum_{k=1}^{K} A^{(k)}\,\ell^{(k)}(\theta)
\;+\; \beta\,\mathrm{KL}\!\left(\pi_{\theta}(\cdot|x)\,\|\,\pi_{\mathrm{ref}}(\cdot|x)\right).
\]
In practice, temperature/top-$p$ controls, sequence-length penalties, entropy scheduling, and rejection of malformed traces stabilize on-policy sampling while preserving the verifiable nature of $r^{(k)}$. Recent research have explored GRPO for video-LLMs, including VideoChat-R1~\cite{li2025videochatr1enhancingspatiotemporalperception}, SpaceR~\cite{ouyang2025spacerreinforcingmllmsvideo}, Fact-R1 (final RL stage)~\cite{zhang2025factr1explainablevideomisinformation}, Reinforcement Learning Tuning for VideoLLMs~\cite{li2025reinforcementlearningtuningvideollms}, Scaling RL to Long Videos~\cite{chen2025scalingrllongvideos}, Video-RTS~\cite{wang2025videortsrethinkingreinforcementlearning}, DeepVideo-R1~\cite{park2025deepvideor1videoreinforcementfinetuning}, Ego-R1 Agent~\cite{tian2025egor1chainoftoolthoughtultralongegocentric}, and so on~\cite{lu2025ovis2,kulkarni2025avatar,ge2025framemind,hong2025glm,wang2025reasoningtrack,he2025framethinker,tao2025moss,guo2025tar,yu2025unhackable,sun2025videosafety,ghazanfari2025chain,fu2025love,yang2025meter,gong2025reinforcing,yang2025kwai,chen2025versavidr1versatilevideounderstanding,long2025adsqa,chen2025chronoforge,li2025tempsamp,team2025kimi,wen2025busterxmllmpoweredaigeneratedvideo,yue2025tempor0videomllmtemporalvideo,lee2025refocusreinforcementguidedframeoptimization}.

\subsection{Video-Specific Policy Optimization}\label{sec:rl:policy}

\paragraph{Policy and trajectory formulation.}
Let $x=(V,q)$ denote the video and query. A trajectory $\tau$ interleaves reasoning and decision tokens,
\[
\tau=\big(r_1,\ldots,r_{k_1},\, d_1,\, r_{k_1+1},\ldots,\, d_2,\ldots,\, y\big),
\]
where decisions may propose temporal spans $[t_s,t_e]$, select keyframes $\mathcal{F}$, emit spatio-temporal regions, and finally produce the answer $y$. The policy $\pi_\theta(\tau\mid x)$ factorizes autoregressively. For a group of $K$ rollouts $\{\tau^{(k)}\}$ with verifiable base rewards $r_{\text{base}}^{(k)}\!\in\![0,1]$, let $\bar r=\tfrac{1}{K}\sum_j r^{(j)}$.

\paragraph{Temporal GRPO (T\text{-}GRPO).}
For each $(V,q)$, construct two input settings: the ordered frame sequence and a randomly shuffled sequence. Generate two groups of responses and compute the proportions of correct answers $p_{\mathrm{ord}}$ and $p_{\mathrm{shuf}}$. Define a temporal coefficient with margin $m\!\ge\!0$:
\[
c_{\text{temp}}=\max\big(0,\;p_{\mathrm{ord}}-p_{\mathrm{shuf}}-m\big).
\]
For an ordered rollout $k$, shape the reward
\[
r^{(k)} \;=\; r_{\text{base}}^{(k)} \;+\; \lambda_{\text{temp}}\, c_{\text{temp}}\;\mathbb{1}\!\big[\text{correct}(\tau^{(k)})\big],
\]
and set the group-relative advantage $A^{(k)}=r^{(k)}-\bar r$. The GRPO update maximizes $\sum_k A^{(k)} \sum_{t\in\tau^{(k)}} \log \pi_\theta(y_t\mid x,y_{<t})$ under a KL anchor to $\pi_{\text{ref}}$, which explicitly rewards accuracy that depends on temporal order rather than single-frame shortcuts~\cite{feng2025videor1reinforcingvideoreasoning}.

\paragraph{Regressive GRPO (Reg-GRPO).}
Reg-GRPO~\cite{park2025deepvideor1videoreinforcementfinetuning} reformulates GRPO as regression on group-normalized advantages, removing min/clipping safeguards. Let the normalized target be
\[
\tilde A^{(k)} \;=\; \frac{r_{\text{base}}^{(k)}-\mu_r}{\sigma_r}, \qquad
\mu_r=\tfrac{1}{K}\sum_j r_{\text{base}}^{(j)},\; \sigma_r=\sqrt{\tfrac{1}{K}\sum_j (r_{\text{base}}^{(j)}-\mu_r)^2}.
\]
Define a sequence score $s_\theta(\tau^{(k)},x)=\sum_{t\in\tau^{(k)}} \log \pi_\theta(y_t\mid x,y_{<t})$. The loss is
\[
\mathcal{L}_{\text{Reg-GRPO}}(\theta)
=\frac{1}{K}\sum_{k=1}^{K}\Big(s_\theta(\tau^{(k)},x)-\tilde A^{(k)}\Big)^2
+\beta\,\mathrm{KL}\!\left(\pi_\theta(\cdot\mid x)\,\|\,\pi_{\text{ref}}(\cdot\mid x)\right).
\]
To mitigate vanishing advantages on very easy/hard samples, DeepVideo-R1 adds difficulty-aware augmentation and/or per-sample weights $w(d(x))$:
\[
\mathcal{L}_{\text{Reg-GRPO}}^{\text{DA}}(\theta)=\frac{1}{K}\sum_{k} w\!\big(d(x)\big)\Big(s_\theta(\tau^{(k)},x)-\tilde A^{(k)}\Big)^2 + \beta\,\mathrm{KL}(\cdot).
\]

\paragraph{Token-weighted advantages (TW-GRPO).}
To improve credit assignment along long chains of thought, TW\text{-}GRPO introduces token importance $w_t$ estimated from intra-group informativeness (\eg, entropy across the $K$ rollouts). Replace the unweighted score with
\[
s_\theta^{\text{TW}}(\tau^{(k)},x)=\sum_{t\in\tau^{(k)}} w_t\,\log \pi_\theta(y_t\mid x,y_{<t}),
\]
and compute advantages from a soft multi-bin reward $r_{\text{soft}}^{(k)}=\sum_b \gamma_b\,\mathbb{1}[y^{(k)}\!\in\!\mathcal{Y}_b]$ (exact, near-miss, wrong). The resulting GRPO/Reg-GRPO objective uses $s_\theta^{\text{TW}}$ in place of $s_\theta$, yielding denser, lower-variance updates~\cite{dang2025reinforcingvideoreasoningfocused}.

\paragraph{Difficulty-aware GRPO (DGRPO).}
To address the difficulty imbalance across tasks or prompts, DGRPO reweights the group-relative advantages by adaptive difficulty signals. Let $d_{\text{task}}$ be a moving hardness estimate at the task level and $d_{\text{sample}}$ a per-prompt score (\eg, running success rate or verifier score dispersion). With a monotone weight $g(\cdot,\cdot)$,
\[
\tilde A_{\text{DA}}^{(k)} \;=\; g\!\big(d_{\text{task}},d_{\text{sample}}\big)\,\big(r^{(k)}-\bar r\big),
\]
and the update maximizes $\sum_k \tilde A_{\text{DA}}^{(k)} \sum_{t\in\tau^{(k)}}\log \pi_\theta(y_t\mid x,y_{<t})$ under the same KL anchor. In “Thinking With Videos,” this scheme is used together with curated multi-task RL data (MTVR-RL-110k) to emphasize informative failures and prevent easy examples from dominating~\cite{zhang2025thinkingvideosmultimodaltoolaugmented}.

\begin{table}[!ht]
\centering
\footnotesize
\setlength{\tabcolsep}{5pt}
\renewcommand{\arraystretch}{1.16}
\caption{Policy optimization methods for Video-LMM post-training. GRPO-family, preference-based alignment, verifier-guided pipelines, and long-video variants.}
\label{tab:grpo_family_3col_final}
\begin{tabularx}{\textwidth}{>{\raggedright\arraybackslash}p{0.15\textwidth} >{\raggedright\arraybackslash}p{0.52\textwidth} >{\raggedright\arraybackslash}p{0.28\textwidth}}
\toprule
\textbf{Method} & \textbf{Objective} & \textbf{Symbols} \\
\midrule
\textbf{Vanilla GRPO}~\cite{li2025reinforcementlearningtuningvideollms,li2025videochatr1enhancingspatiotemporalperception} &
$\displaystyle
\max_{\theta}\frac{1}{G}\sum_{i=1}^{G}
\min\!\Big(\tfrac{\pi_{\theta}(y_i)}{\pi_{\text{old}}(y_i)}A_i,\,
\mathrm{clip}\big(\tfrac{\pi_{\theta}}{\pi_{\text{old}}},1-\epsilon,1+\epsilon\big)A_i\Big)
-\beta\,\mathrm{KL}\!\big[\pi_{\theta}\,\|\,\pi_{\text{ref}}\big]
$ &
$G$: group size; $A_i=\frac{r_i-\mu_r}{\sigma_r}$; $r_i$: verifiable reward; $\pi_{\text{ref}}$: reference policy; $\mathrm{clip}(\cdot)$: PPO-style clipping \\
\midrule
\textbf{T-GRPO}~\cite{feng2025videor1reinforcingvideoreasoning} &
$\displaystyle
\max_{\theta}\,\mathcal{L}_{\text{GRPO}}(\theta)\;+\;\lambda_t\,\alpha\,\mathbf{1}\!\big[p_{\mathrm{ord}}>p_{\mathrm{shuf}}\big]
$ &
$p_{\mathrm{ord}},p_{\mathrm{shuf}}$: success on ordered vs. shuffled frames; $\lambda_t,\alpha$: weights \\
\midrule
\textbf{TW-GRPO}~\cite{dang2025reinforcingvideoreasoningfocused} &
$\displaystyle
\max_{\theta}\frac{1}{G}\sum_i
\min\!\Big(\tfrac{\pi_{\theta}}{\pi_{\text{old}}}A'_i,\mathrm{clip}(\cdot)A'_i\Big)
-\beta\,\mathrm{KL},\quad
A'_i=\sum_{t} w_t\,a_{it},\quad
r=\sum_{k}\gamma_k\,\mathbf{1}[y\in \mathcal{Y}_k]
$ &
$A'_i$: token-weighted advantage; $w_t$: token importance; $a_{it}$: token-level advantage; $\mathcal{Y}_k$: partial-credit bins; $\gamma_k$: bin weights \\
\midrule
\textbf{Reg-GRPO}~\cite{park2025deepvideor1videoreinforcementfinetuning} &
$\displaystyle
\min_{\theta}\;\frac{1}{G}\sum_{i=1}^{G}
\Big(\Delta\log\pi_{\theta}(y_i)-\eta A_i\Big)^2
\;+\;\beta\,\mathrm{KL}
$ &
$\Delta\log\pi_{\theta}(y_i)=\log\!\tfrac{\pi_{\theta}(y_i)}{\pi_{\text{old}}(y_i)}$; $\eta$: regression scale \\
\midrule
\textbf{DGRPO}~\cite{zhang2025factr1explainablevideomisinformation} &
$\displaystyle
\max_{\theta}\frac{1}{G}\sum_i
\min\!\Big(\tfrac{\pi_{\theta}}{\pi_{\text{old}}}A_i'',\mathrm{clip}(\cdot)A_i''\Big)
-\beta\,\mathrm{KL},\quad
A_i'' = w\!\big(d(x)\big)\cdot A_i
$ &
$d(x)$: difficulty score; $w(\cdot)$: difficulty weight; $A_i''$: difficulty-weighted advantage \\
\midrule
\textbf{Multi-task GRPO}~\cite{li2025videochatr1enhancingspatiotemporalperception} &
$\displaystyle
\max_{\theta}\frac{1}{G}\sum_i
\min\!\Big(\tfrac{\pi_{\theta}}{\pi_{\text{old}}}\textstyle\sum_m \lambda_m A_i^{(m)},\,
\mathrm{clip}(\cdot)\sum_m \lambda_m A_i^{(m)}\Big)
-\beta\,\mathrm{KL}
$ &
$A_i^{(m)}$: standardized advantage on task $m$; $\lambda_m$: task weights \\
\midrule
\textbf{Verifier-DPO}~\cite{li2025veripocultivatinglongreasoning} &
$\displaystyle
\min_{\theta}\; \mathcal{L}_{\text{DPO}}
=-\log\frac{\exp(\beta s^{+})}{\exp(\beta s^{+})+\exp(\beta s^{-})}
$ &
$s^{+},s^{-}$: scores of preferred/rejected outputs; $\beta$: DPO temperature \\
\midrule
\textbf{Long-video-RL}~\cite{chen2025scalingrllongvideos} &
$\displaystyle
\max_{\theta}\; \sum_{s=1}^{S}\mathcal{L}_{\text{GRPO}}^{(s)}(\theta)\;-\;\gamma\,\Omega(\text{memory/retrieval})
$ &
$S$: \#segments; $\mathcal{L}_{\text{GRPO}}^{(s)}$: per-segment objective; $\Omega(\cdot)$: memory/retrieval regularizer; $\gamma$: weight \\
\midrule
\textbf{Temporal-only grounding RL }~\cite{li2025reinforcementlearningtuningvideollms} &
$\displaystyle
\max_{\theta}\;\mathcal{L}_{\text{GRPO}}(\theta)
\quad\text{s.t.}\quad r=\mathrm{IoU}\!\big([t_s,t_e],\hat{[t_s,t_e]}\big)
$ &
$[t_s,t_e]$: predicted span; $\hat{[t_s,t_e]}$: ground-truth span \\
\midrule
\textbf{Spatio-temporal GRPO}~\cite{li2025videochatr1enhancingspatiotemporalperception} &
$\displaystyle
\max_{\theta}\;\mathcal{L}_{\text{GRPO}}(\theta)
\quad\text{with}\quad
r=\lambda_f R_{\text{format}}+\lambda_{\mathrm{IoU}} R_{\mathrm{IoU}}+\lambda_{a} R_{\text{acc}}+\lambda_{r} R_{\text{recall}}
$ &
$R_{\text{format}}$: structured output; $R_{\mathrm{IoU}}$: temporal IoU; $R_{\text{acc}}$: MC/classification accuracy; $R_{\text{recall}}$: event recall \\
\midrule
\textbf{Caption w/ dual verifiable rewards}~\cite{meng2025videocapr1enhancingmllmsvideo} &
$\displaystyle
\max_{\theta}\;\mathcal{L}_{\text{GRPO}}(\theta)
\quad\text{with}\quad
r=\lambda_f R_{\text{format}}+\lambda_c R_{\text{content}}
$ &
$R_{\text{format}}$: template/structure score; $R_{\text{content}}$: content fidelity; $\lambda_f,\lambda_c$: weights \\
\midrule
\textbf{RL$\times$RTS}~\cite{wang2025videortsrethinkingreinforcementlearning} &
$\displaystyle
\max_{\theta}\;\mathcal{L}_{\text{GRPO}}(\theta)\;-\;\lambda_s\,\Phi(\text{CoT steps})
$ &
$\lambda_s$: coupling weight; $\Phi(\cdot)$: penalty/constraint on CoT step count \\
\bottomrule
\multicolumn{3}{l}{\textit{Note: DPO can be viewed as an offline preference-based alignment method related to RL.}}
\end{tabularx}
\end{table}

\begin{table}[!ht]
\centering
\footnotesize
\setlength{\tabcolsep}{5pt}
\renewcommand{\arraystretch}{1.16}
\caption{Reward design taxonomy for Video-LMM post-training.}
\label{tab:reward_design_video_llm}
\begin{tabularx}{\textwidth}{>{\raggedright\arraybackslash}p{0.20\textwidth} >{\raggedright\arraybackslash}p{0.52\textwidth} >{\raggedright\arraybackslash}p{0.24\textwidth}}
\toprule
\textbf{Aspect} & \textbf{Typical formulation} & \textbf{Examples} \\
\midrule
Temporal localization & Span IoU/mIoU; event order consistency; count/duration constraints & \cite{li2025reinforcementlearningtuningvideollms,li2025videochatr1enhancingspatiotemporalperception,feng2025videor1reinforcingvideoreasoning} \\
\midrule
Spatial grounding & Box/mask/track IoU; trajectory overlap; relation/pose consistency & \cite{li2025videochatr1enhancingspatiotemporalperception,ouyang2025spacerreinforcingmllmsvideo} \\
\midrule
Content correctness & MC accuracy; open-ended semantic match; partial-credit bins & \cite{dang2025reinforcingvideoreasoningfocused,zhang2025factr1explainablevideomisinformation} \\
\midrule
Format/structure & Enforce \texttt{<think>/<answer>} template; reasoning-step completeness & \cite{zhang2025factr1explainablevideomisinformation,li2025reinforcementlearningtuningvideollms} \\
\midrule
Hallucination mitigation & Entity/evidence grounding checks; cross-modal consistency penalty & \cite{gao2025exploringhallucinationlargemultimodal,li2025vidhallucevaluatingtemporalhallucinations} \\
\midrule
Difficulty-aware weighting & $w(d(x))$ on advantages; curriculum by hardness bins & \cite{zhang2025thinkingvideosmultimodaltoolaugmented} \\
\midrule
Tool-augmented signals & Reward for informative frame retrieval; toolbox success/failure & \cite{zhang2025thinkingvideosmultimodaltoolaugmented,tian2025egor1chainoftoolthoughtultralongegocentric} \\
\midrule
Memory/retrieval regularization & Penalty $\Omega(\cdot)$ on memory calls; segment-wise consistency & \cite{chen2025scalingrllongvideos} \\
\midrule
Audio-aware consistency & Optional ASR/AV alignment scores when audio is used & \cite{sun2025videosalmonno1reasoningenhancedaudiovisuallarge} \\
\bottomrule
\end{tabularx}
\end{table}

\subsection{Reward Design for Video Reasoning}\label{sec:rl:rewards}
We decompose the outcome reward into verifiable components and aggregate them with task weights:
\[
R(x,\tau)=\sum_{m}\lambda_m\,R_m(x,\tau),\qquad \lambda_m\ge 0,\;\sum_m\lambda_m=1,
\]
which distributes incentives and mitigates reward hacking by avoiding reliance on any single objective.

\paragraph{Format and faithfulness.}
Outputs are parsed with lightweight rules (\eg, required <think>/<answer> tags, unit normalization, timestamp presence, citation syntax). Violations incur graded penalties; contradictions with visual or subtitle evidence trigger additional deductions~\cite{li2025veripocultivatinglongreasoning}.

\paragraph{Answer correctness.}
For multiple-choice, we use exact match. For open-ended responses, we compute normalized string scores (\eg, edit distance, token-F1) with minor lexical normalization and, when necessary, a calibrated evaluator to assign partial credit rather than binary pass/fail~\cite{wang2025videorftincentivizingvideoreasoning}.

\paragraph{Temporal localization.}
Given a predicted interval $P=[t_s,t_e]$ and ground truth $G$, we combine smooth temporal IoU and threshold bonuses while discouraging overlong spans:
\[
R_{\text{temp}}=\alpha\,\mathrm{tIoU}(P,G)+\sum_{k}b_k\,\mathbb{1}\!\left[\mathrm{tIoU}(P,G)\ge \tau_k\right]-\gamma\,\tfrac{|P|}{|V|}.
\]
Missed critical events (false negatives) receive additional penalties to avoid degenerate short spans~\cite{xu2025viarladaptivetemporalgrounding,luo2025musegreinforcingvideotemporal}.

\paragraph{Spatio-temporal grounding.}
For regions or tracks $\{B_t\}$, we combine region-IoU/track-IoU with center-distance shaping and enforce text–region referential consistency across frames to prevent hallucinated references~\cite{li2025videochatr1enhancingspatiotemporalperception}.

\paragraph{Budget awareness.}
Let $B$ be the frame/token budget. We reward accurate solutions that respect $B$ and penalize redundant re-observations; staged viewing (coarse-to-fine frame selection) receives a small bonus:
\[
R_{\text{budget}}=\eta_1\,\mathbb{1}[\text{correct}]\cdot\Big(1-\tfrac{\text{used}}{B}\Big)-\eta_2\,\tfrac{\text{repeats}}{\text{used}}.
\]
This keeps the policy sample-efficient during long-video rollouts~\cite{jiang2025stormtokenefficientlongvideo,hu2025mllmbasedvideoframe}.

\paragraph{Verifier and critic signals.}
External verifiers check timestamp/region claims and entity references; multi-path self-consistency (\eg, majority vote or agreement rate across $K$ sampled traces) yields pass/fail or graded signals that fold into $R$ and help stabilize exploration~\cite{li2025veripocultivatinglongreasoning,zhang2025thinkingvideosmultimodaltoolaugmented}.

\paragraph{Aggregation and normalization.}
Task weights $\{\lambda_m\}$ are tuned to equalize gradient magnitudes across objectives. We normalize each $R_m$ to $[0,1]$ on a per-batch basis and apply temperature scaling when mixing discrete pass/fail terms with continuous IoU-style signals. This keeps the GRPO advantages well-conditioned and reduces variance during on-policy sampling.

\subsection{RL Datasets for Video Reasoning}\label{sec:rl:data}
Reinforcement learning for video reasoning draws on three complementary data sources. First, supervised chain-of-thought corpora warm up the policy to produce structured traces that can be scored online by verifiers. Second, RL rollout corpora provide prompts with verifiable targets, \eg, answer strings, timestamps, or regions, so that outcome rewards can be computed without human preferences. Third, curated hard negatives and near-duplicate distractors sharpen temporal and spatial discrimination under limited budgets.

\paragraph{Representative scales and staging.}
Across recent Video-LMMs the RL data footprint ranges from a few thousand to hundreds of thousands of examples, often after a smaller SFT warmup. Video-RTS demonstrates a single-stage GRPO pipeline trained on roughly $6$K video–QA triples, yet matches systems that rely on $\sim\!165$K SFT pairs, highlighting data efficiency under verifiable rewards~\cite{wang2025videortsrethinkingreinforcementlearning}. LongVILA adopts a two-phase schedule: long-video CoT-SFT on about $36$K samples, followed by GRPO with $\sim\!68$K filtered prompts plus $\sim\!102$K external additions to stabilize exploration at length~\cite{chen2025scalingrllongvideos}. Fact-R1 explicitly separates stages, $\sim\!85$K long-form CoT-SFT, then $\sim\!5$K preference pairs for DPO alignment, and finally GRPO with verifier-backed outcome rewards while jointly training auxiliary caption/OCR heads~\cite{zhang2025factr1explainablevideomisinformation}. Multi-task GRPO in VideoChat-R1 operates over a mixed training set totaling approximately $18{,}031$ samples spanning QA, grounding, tracking, and captioning, showing that a moderate-scale, heterogeneous pool suffices when rewards are verifiable~\cite{li2025videochatr1enhancingspatiotemporalperception}. Larger pipelines exist as well: ARC-Hunyuan-Video-7B~\cite{ge2025archunyuanvideo7bstructuredvideocomprehension} reports instruction-tuning corpora on the order of $4.6\times10^5$ pairs and tens of thousands of GRPO rollouts distributed across tasks, interleaved with cold-start and polish stages to control drift.

\paragraph{Temporal and spatial supervision.}
Effective RL corpora emphasize prompts with temporal anchors and spatial references so that rewards can combine correctness with localization. Typical sources include timestamped QA, dense event or action segments, and region-grounded queries. For long-form content, authors construct silver labels with shot detection and ASR alignment to produce answerable windows and span-level targets, which enable smooth tIoU shaping during GRPO~\cite{chen2025scalingrllongvideos,wang2025videortsrethinkingreinforcementlearning}.

\paragraph{Curation and filtering.}
To control reward hacking and variance, recent works filter prompts for unambiguous answers, enforce strict formatting constraints, and mine hard negatives from near-duplicate shots or distractor spans before rollout. In practice this yields a compact but high-yield RL pool (\eg, the $\sim\!68$K filtered set in LongVILA) that keeps the verifier precise and the advantages well-conditioned~\cite{chen2025scalingrllongvideos}.

\paragraph{Domain breadth and streaming settings.}
Beyond general video QA, RL datasets extend to navigation, egocentric, and streaming regimes where budgets and latency matter. For example, StreamVLN trains over hundreds of thousands of trajectories and on the order of $6\times10^7$ frames with a GRPO-style objective adapted to streaming perception and action, illustrating how outcome rewards transfer to embodied video tasks~\cite{wei2025streamvlnstreamingvisionandlanguagenavigation}.

\section{Test-Time Scaling for Video Reasoning}\label{sec:tts}

\begin{myboxi}[Takeaways]
Test-time scaling improves reliability by allocating inference compute to evidence selection, reasoning depth, and path diversity. Recent work has explored various TTS strategies, including Video-CoT prompting, self-consistency with verifier gating, confidence-guided iteration with refine-on-fail, and tool-augmented chains for long or streaming videos.
\end{myboxi}

\subsection{Beam Search for Video Outputs}
Beam search is a standard decoding strategy adopted by many video captioning and video-QA models to improve the fluency and relevance of generated text. In video captioning tasks, for example, models often generate descriptions using beam search (e.g., beam width 5) to explore multiple candidate sentences and pick the best one. This approach has been used to produce higher-quality captions by balancing completeness and coherence as compared to greedy decoding. Overall, beam search serves as a test-time decoding boost for Video-LMMs by considering alternative word sequences and selecting the highest-probability caption.

\subsection{Video Chain-of-Thought Prompting}
CoT prompting, getting the model to generate intermediate reasoning steps before the final answer, has been successfully extended to video understanding. Video-of-Thought (VoT)~\cite{fei2024videoofthoughtstepbystepvideoreasoning} was one of the first frameworks to implement CoT for video reasoning. VoT~\cite{fei2024videoofthoughtstepbystepvideoreasoning} breaks a complex video question into simpler sub-problems and addresses them step by step, from low-level perceptual cues to high-level conclusions. This explicit reasoning significantly improved performance on challenging video QA benchmarks, demonstrating the benefit of prompted reasoning traces in video tasks. More recently, CoT-Vid~\cite{jin2025cotviddynamicchainofthoughtrouting} introduced a training-free multi-stage CoT pipeline for video QA. CoT-Vid~\cite{jin2025cotviddynamicchainofthoughtrouting} dynamically decides whether a question needs reasoning, then decomposes it and iteratively reasons step by step before producing the answer, yielding notable accuracy gains without any model fine-tuning.

\subsection{Self-Consistency Decoding in Video Reasoning}
Video-LMMs have also begun to employ self-consistency decoding, where multiple reasoning paths are sampled and then aggregated to improve answer reliability. A clear example is the video self-consistency verification stage in CoT-Vid~\cite{jin2025cotviddynamicchainofthoughtrouting}. During inference, CoT-Vid~\cite{jin2025cotviddynamicchainofthoughtrouting} generates multiple reasoning chains for the same question and uses a similarity-based voting mechanism to merge them into a final answer. This ensures that the chosen answer is consistent with the majority of reasoning paths and with the video content, reducing random errors or hallucinations. Empirically, video self-consistency yields better accuracy as more answer samples are considered, CoT-Vid’s performance improved steadily up to about five reasoning samples before saturating, stabilizing outputs by leveraging ensemble reasoning.

\subsection{Confidence-Based Iterative Reasoning}
Recent Video-LMM agents use confidence measures to guide and terminate multi-step inference. CyberV~\cite{meng2025cybervcyberneticstesttimescaling} treats reasoning as a closed-loop process: a controller monitors uncertainty and instructs the model to think deeper or request denser visual evidence until a stopping criterion is met. Video-ICL~\cite{kim2024videoiclconfidencebasediterativeincontext} similarly allocates more computation to uncertain queries and stops early on confident ones. This confidence-driven iteration allows Video-LMMs to balance thoroughness and efficiency by refining their understanding progressively and stopping only when the answer is likely correct.

\subsection{Self-Improvement via Refinement Loops}
Several video reasoning frameworks implement iterative self-refinement loops at test time, enabling the model to improve answers over multiple rounds. DIVE (Deep-search Iterative Video Exploration)~\cite{kamoto2025divedeepsearchiterativevideo} breaks down each question into sub-questions and tackles them in a multi-step loop, refining the queries and answers at each pass. If an intermediate answer is incomplete or a sub-question remains, DIVE~\cite{kamoto2025divedeepsearchiterativevideo} re-evaluates and refines that part in the next iteration. This refine-on-fail strategy yields highly accurate and contextually appropriate answers even for complex queries. Similarly, Video-MTR~\cite{xie2025videomtrreinforcedmultiturnreasoning} performs multi-turn reasoning on long videos, progressively selecting relevant segments and updating the answer until convergence.

\subsection{Monte Carlo Tree Search (MCTS) for Video-LMMs}
Monte Carlo Tree Search has been applied to expand and diversify generation at inference. AutoCaption~\cite{yu2025evaluatingmllmsvideocaptioningmcts} uses MCTS to iteratively construct diverse video descriptions by exploring a tree of possible continuations and selecting branches that yield informative sentences. This produces rich sets of key-point captions that go beyond fixed-beam decoding, and enables the MCTS-VCB benchmark where MLLMs fine-tuned on AutoCaption outputs show large gains.

\subsection{Chain-of-Action and Tool-Augmented Reasoning}
Video-LMMs are increasingly embracing tool use and multi-step action chains to handle complex video understanding. VITAL~\cite{zhang2025thinkingvideosmultimodaltoolaugmented} equips a video-language model with a visual toolbox that the model can call during reasoning. At inference time, VITAL~\cite{zhang2025thinkingvideosmultimodaltoolaugmented} decides when to invoke tools (for example, to fetch a particular video clip segment or detect an object) and incorporates the results into a multimodal chain of thought, greatly reducing hallucinations by grounding intermediate claims in returned evidence. Ego-R1~\cite{tian2025egor1chainoftoolthoughtultralongegocentric} introduces a Chain-of-Tool-Thought paradigm for ultra-long egocentric videos: an RL-trained agent orchestrates specialized tools in sequence, first calling a temporal retrieval tool to find a relevant moment, then an object recognizer, and so on, each tool tackling a sub-task of the query, enabling answers about weeks-long recordings beyond raw context limits. ReAgent-V~\cite{zhou2025reagentvrewarddrivenmultiagentframework} coordinates multiple specialized agents and tools so that perception and reasoning are scheduled and verified under long or streaming inputs. Complementary agentic strategies include VideoDeepResearch~\cite{yuan2025videodeepresearchlongvideounderstanding}, which performs tool-augmented search over long videos at inference time, and Agentic Keyframe Search~\cite{fan2025agentickeyframesearchvideo}, which plans which frames to inspect and couples planner–executor loops with verification before answer commitment.

\begin{table}[!ht]
\centering
\footnotesize
\setlength{\tabcolsep}{5pt}
\renewcommand{\arraystretch}{1.12}
\caption{Datasets used in Video-LLM post-training (training \& evaluation). Row color indicates primary usage scenario, and datasets may be used across multiple stages: \colorbox{SFTColor}{SFT}, \colorbox{RLColor}{RL}, \colorbox{BenchColor}{Bench}.}
\label{tab:data_summary_pt}
\resizebox{0.83\linewidth}{!}{%
\begin{tabularx}{\textwidth}{
  >{\raggedright\arraybackslash}p{0.25\textwidth}
  >{\raggedright\arraybackslash}p{0.18\textwidth}
  >{\raggedright\arraybackslash}X
  >{\centering\arraybackslash}p{0.08\textwidth}
}
\toprule
\textbf{Name (with source)} & \textbf{Size} & \textbf{Tasks} & \textbf{Link} \\
\midrule
\rowcolor{SFTColor}
Temporal-RLT-Full-490k~\cite{li2025reinforcementlearningtuningvideollms} & 490{,}000 & VideoQA, temporal grounding, grounded VideoQA; diversified difficulty; used before RL. & \href{https://huggingface.co/datasets/appletea2333/temporal_r1}{\faHuggingFace} \\
\midrule
\rowcolor{RLColor}
Temporal-RLT-32k~\cite{li2025reinforcementlearningtuningvideollms} & 32{,}000 & Curated subset for GRPO-style RLT; temporal signals emphasized. & \href{https://huggingface.co/datasets/appletea2333/temporal_r1}{\faHuggingFace} \\
\midrule
\rowcolor{SFTColor}
VideoChat-R1 training set~\cite{li2025videochatr1enhancingspatiotemporalperception} & 18{,}031 & Multi-task SFT covering grounding, tracking, grounded QA. & -- \\
\midrule
\rowcolor{SFTColor}
MTVR-CoT-72k~\cite{zhang2025thinkingvideosmultimodaltoolaugmented} & 72{,}000 & Long CoT reasoning; temporal grounding; tool-augmented SFT variants included. & \href{https://huggingface.co/datasets/zhang9302002/MultiTaskVideoReasoning}{\faHuggingFace} \\
\midrule
\rowcolor{RLColor}
MTVR-RL-110k~\cite{zhang2025thinkingvideosmultimodaltoolaugmented} & 110{,}000 & Multi-task video reasoning; difficulty-aware scheduling. & \href{https://huggingface.co/datasets/zhang9302002/MultiTaskVideoReasoning}{\faHuggingFace} \\
\midrule
\rowcolor{SFTColor}
Video-R1-COT-165k~\cite{feng2025videor1reinforcingvideoreasoning} & 165{,}000 & Chain-of-thought supervision for time-aware reasoning (ordered vs. shuffled frames). & \href{https://huggingface.co/datasets/Video-R1/Video-R1-data}{\faHuggingFace} \\
\midrule
\rowcolor{RLColor}
Video-R1-260k~\cite{feng2025videor1reinforcingvideoreasoning} & 260{,}000 & RL pool for T-GRPO reinforcement; mixed video/image subsets. & \href{https://huggingface.co/datasets/Video-R1/Video-R1-data}{\faHuggingFace} \\
\midrule
\rowcolor{SFTColor}
video-SALMONN-o1 (QA pairs)~\cite{sun2025videosalmonno1reasoningenhancedaudiovisuallarge} & $\sim$180{,}000 QA (from $\sim$13k videos) & Audio+video reasoning; curated QA pairs for instruction/SFT. & -- \\
\midrule
\rowcolor{RLColor}
video-SALMONN-o1 (preferences)~\cite{sun2025videosalmonno1reasoningenhancedaudiovisuallarge} & $\sim$200{,}000 pairs & Pairwise preference data for DPO/RFT-like objectives; strengthens chain-of-thought quality. & -- \\
\midrule
\rowcolor{SFTColor}
LongVILA CoT-SFT~\cite{chen2025scalingrllongvideos} & 36{,}000 & Long-video chain-of-thought supervision; length-aware prompts. & \href{https://huggingface.co/datasets/LongVideo-Reason/longvideo-reason}{\faHuggingFace} \\
\midrule
\rowcolor{RLColor}
LongVILA RL pool~\cite{chen2025scalingrllongvideos} & 68{,}000 + 102{,}000 (open) & Two-part RL data (in-house + open-source) targeting long temporal reasoning. & \href{https://huggingface.co/datasets/LongVideo-Reason/longvideo-reason}{\faHuggingFace} \\
\midrule
\rowcolor{SFTColor}
FakeVV (news-domain)~\cite{zhang2025factr1explainablevideomisinformation} & 197{,}600 & Video misinformation detection/explanation; reasoning traces. & \href{https://github.com/zfr00/fact-r1}{\faGithub} \\
\midrule
\rowcolor{SFTColor}
FakeTT (short-video, EN)~\cite{zhang2025factr1explainablevideomisinformation} & — & Short-video misinformation (English); used for SFT and analysis. & \href{https://github.com/zfr00/fact-r1}{\faGithub} \\
\midrule
\rowcolor{SFTColor}
FakeSV (short-video, ZH)~\cite{zhang2025factr1explainablevideomisinformation} & 18{,}859 & Short-video misinformation (Chinese); reasoning. & \href{https://github.com/zfr00/fact-r1}{\faGithub} \\
\midrule
\rowcolor{SFTColor}
TVG-Coldstart-13K~\cite{chen2025datasets} & $\sim$13k & SFT cold-start for temporal grounding & \href{https://huggingface.co/datasets/RuizheChen/TVG_processed_data}{\faHuggingFace} \\
\midrule
\rowcolor{RLColor}
TVG-RL-18K~\cite{chen2025datasets} & $\sim$18k & RL data for temporal grounding & \href{https://huggingface.co/datasets/RuizheChen/TVG_processed_data}{\faHuggingFace} \\
\midrule
\rowcolor{BenchColor}
Charades-STA~\cite{li2025videochatr1enhancingspatiotemporalperception} & — & Temporal grounding benchmark. & \href{https://huggingface.co/datasets/VLM2Vec/Charades-STA}{\faHuggingFace} \\
\midrule
\rowcolor{BenchColor}
ActivityNet-Grounding~\cite{li2025videochatr1enhancingspatiotemporalperception} & — & Temporal grounding benchmark. & \href{https://github.com/facebookresearch/ActivityNet-Entities}{\faGithub} \\
\midrule
\rowcolor{BenchColor}
ActivityNet-RTL~\cite{li2025reinforcementlearningtuningvideollms,li2025videochatr1enhancingspatiotemporalperception} & — & Reasoning-intensive temporal grounding benchmark. & \href{https://huggingface.co/datasets/deahuang/LITA-Datasets/tree/main}{\faHuggingFace} \\
\midrule
\rowcolor{BenchColor}
AVE-2~\cite{vosoughi2025ave2} & 570{,}138 & Audio-visual alignment evaluation reasoning. & \href{https://huggingface.co/datasets/ali-vosoughi/ave-2}{\faHuggingFace} \\
\midrule
\rowcolor{BenchColor}
GoT--10k~\cite{li2025videochatr1enhancingspatiotemporalperception} & — & Object tracking benchmark. & \href{https://huggingface.co/huangyuyang11/got10k}{\faHuggingFace} \\
\midrule
\rowcolor{BenchColor}
NExT-GQA~\cite{li2025videochatr1enhancingspatiotemporalperception} & — & Video QA / grounded QA benchmark. & \href{https://huggingface.co/datasets/jinyoungkim/NExT-GQA}{\faHuggingFace} \\
\midrule
\rowcolor{BenchColor}
Dream--1k~\cite{li2025videochatr1enhancingspatiotemporalperception} & — & Captioning benchmark (dense descriptions). & \href{https://huggingface.co/datasets/omni-research/DREAM-1K}{\faHuggingFace} \\
\midrule
\rowcolor{BenchColor}
VidTAB-QA~\cite{li2025videochatr1enhancingspatiotemporalperception} & — & Video QA quality assessment benchmark. & \href{https://github.com/MCG-NJU/VideoEval}{\faHuggingFace} \\
\midrule
\rowcolor{BenchColor}
VSI-Bench~\cite{ouyang2025spacerreinforcingmllmsvideo} & — & Spatial reasoning (relations, order, counting). & \href{https://huggingface.co/datasets/nyu-visionx/VSI-Bench}{\faHuggingFace} \\
\midrule
\rowcolor{BenchColor}
VideoMME~\cite{li2025videochatr1enhancingspatiotemporalperception} & 2{,}700 QA & General video understanding benchmark. & \href{https://huggingface.co/datasets/lmms-lab/Video-MME}{\faHuggingFace} \\
\midrule
\rowcolor{BenchColor}
MVBench~\cite{li2025videochatr1enhancingspatiotemporalperception} & — & General video understanding benchmark. & \href{https://huggingface.co/datasets/OpenGVLab/MVBench}{\faHuggingFace} \\
\midrule
\rowcolor{BenchColor}
Video-Holmes~\cite{cheng2025videoholmesmllmthinklike} & — & Video reasoning benchmark. & \href{https://huggingface.co/datasets/TencentARC/Video-Holmes}{\faHuggingFace} \\
\midrule
\rowcolor{BenchColor}
MMVU~\cite{zhao2025mmvumeasuringexpertlevelmultidiscipline} & 3{,}000 items & Expert-level multidisciplinary video. & \href{https://huggingface.co/datasets/yale-nlp/MMVU}{\faHuggingFace} \\
\midrule
\rowcolor{BenchColor}
Video-MMMU~\cite{hu2025videommmu} & 900 QA pairs & Multi-discipline professional videos. & \href{https://huggingface.co/datasets/lmms-lab/VideoMMMU}{\faHuggingFace} \\
\midrule
\rowcolor{BenchColor}
VideoHallucer / HAVEN~\cite{gao2025exploringhallucinationlargemultimodal} & 6{,}497 QA (HAVEN) & Hallucination evaluation (object/temporal consistency). & \href{https://huggingface.co/datasets/bigai-nlco/VideoHallucer}{\faHuggingFace} \\
\bottomrule
\end{tabularx}
}
\end{table}
\section{Benchmarks for Video-LMM Post-training Evaluation}\label{sec:benchmarks}

Evaluating post-training requires benchmarks aligned with optimization objectives: verifiable supervision for RL, realistic compute budgets for TTS, and protocols that expose genuine reasoning rather than shortcut exploitation. We organize resources into general QA, video reasoning, and grounding-centric benchmarks, emphasizing settings that enable verifier-ready rewards and standardized comparisons. Table~\ref{tab:data_summary_pt} summarizes commonly used datasets in recent post-training work.

\begin{myboxi}[Takeaways]
Alignment between evaluation metrics and training objectives enables more interpretable optimization: answer faithfulness, temporal correctness, and spatial–temporal grounding under realistic budgets with verifier-ready annotations. The field has moved beyond monolithic QA suites toward targeted evaluations, including multi-event reasoning, long-video and streaming, and precise grounding, that better diagnose where post-training gains come from.
\end{myboxi}

\subsection{General Video QA Benchmarks}

Comprehensive QA suites probe recognition, reasoning, and instruction following across diverse lengths and domains~\cite{fu2025videomme,mangalam2023egoschema,hu2025videommmu}. MMVU~\cite{zhao2025mmvumeasuringexpertlevelmultidiscipline} targets expert-level, multi-discipline understanding and provides dual reporting protocols (with and without subtitles) to expose text-based shortcuts. VCR-Bench~\cite{qi2025vcrbenchcomprehensiveevaluationframework} focuses on compositional, causal, and multi-step reasoning with fine-grained categories for capability analysis. VideoReasonBench~\cite{liu2025videoreasonbenchmllmsperformvisioncentric} emphasizes vision-centric reasoning beyond frame-level recognition, stressing cross-event inference and temporal dependencies. MINERVA~\cite{nagrani2025minervaevaluatingcomplexvideo} stresses complex multi-step reasoning over long videos, assessing sustained attention and multi-hop inference. Standard metrics include accuracy for multiple-choice and exact match or F1 for free-form answers, with recommended dual reporting with and without subtitles to reveal linguistic shortcut exploitation~\cite{zhao2025mmvumeasuringexpertlevelmultidiscipline, qi2025vcrbenchcomprehensiveevaluationframework}.

\subsection{Video Reasoning Benchmarks}

Reasoning-centric evaluations isolate capabilities that post-training often targets. MECD~\cite{chen2024mecdunlockingmultieventcausal} measures multi-event causal dependencies, enabling analysis of causal chains across shots. VidHalluc~\cite{li2025vidhallucevaluatingtemporalhallucinations} and HAVEN~\cite{gao2025exploringhallucinationlargemultimodal} probe hallucination robustness, including temporal hallucination and object consistency, testing whether models fabricate non-existent entities or events. Long-video and streaming settings such as LongVideo-Reason-eval~\cite{chen2025scalingrllongvideos} and streaming/multi-round evaluations (e.g., StreamBench~\cite{xiong2025streamingvideounderstandingmultiround}, SVBench~\cite{yang2025svbenchbenchmarktemporalmultiturn}, OmniMMI~\cite{wang2025omnimmicomprehensivemultimodalinteraction}) stress memory management, budgeted viewing, and stability under temporal resampling. For these protocols, budget- and latency-aware reporting is essential: disclose viewing budget (frames or tokens), reasoning length, path count, and latency/throughput alongside accuracy to reveal cost–performance trade-offs critical for deployment~\cite{chen2025scalingrllongvideos}.

\subsection{Grounding Reasoning Benchmarks for Video-LMMs}

Grounding-centric benchmarks align tightly with verifiable rewards used in RL and with inference-time verification. Temporal localization datasets such as Charades-STA and ActivityNet Grounding~\cite{li2025videochatr1enhancingspatiotemporalperception} evaluate precise moment retrieval from language, while ActivityNet-RTL~\cite{li2025reinforcementlearningtuningvideollms, li2025videochatr1enhancingspatiotemporalperception} requires multi-step reasoning before localization. The fine-grained 0–10 scores make it a verifier-ready resource for
RL-based post-training and a bridge between moment-localization benchmarks
and multimodal reasoning suites. Spatial–temporal grounding benchmarks broaden the target to regions and tracks: V-STAR~\cite{cheng2025vstarbenchmarkingvideollmsvideo} provides entity/action grounding with trajectory annotations; VSI-Bench~\cite{ouyang2025spacerreinforcingmllmsvideo} probes spatial relations, ordering, and counting; GoT-10k~\cite{huang2019got} stresses long-term identity maintenance via object tracking. Evaluation commonly reports temporal IoU (tIoU), Recall@K at multiple tIoU thresholds (e.g., 0.3/0.5/0.7), region/trajectory IoU, and center-distance errors, with locate-then-answer protocols that require models to commit to evidence before producing answers~\cite{xu2025viarladaptivetemporalgrounding, luo2025musegreinforcingvideotemporal, li2025videochatr1enhancingspatiotemporalperception，yuan2025momentseeker}.

\subsection{Long and Streaming Video Evaluation}

Long/streaming evaluations target long-horizon reasoning, dialogue coherence, and timestamp sensitivity under online constraints. {SVBench}~\cite{yang2025svbenchbenchmarktemporalmultiturn} uses temporally linked multi-turn QA chains to probe streaming understanding; {StreamBench}~\cite{xiong2025streamingvideounderstandingmultiround} evaluates real-time, interactive scenarios. {OVO-Bench}~\cite{li2025ovobenchfarvideollmsrealworld} stresses timestamp-aware online reasoning with three settings, backward tracing, real-time comprehension, and forward (delayed) answering, paired with fine-grained temporal annotations. For long-form video, {LongViTU}~\cite{wu2025longvituinstructiontuninglongform} supplies large-scale long-video QA with explicit timestamps, and {HLV-1K}~\cite{zou2025hlv1klargescalehourlongvideo} focuses on hour-long videos. For captioning, AuroraCap~\cite{chai2024auroracap} introduces VDC (a detailed video captioning benchmark) and VDCscore, an LLM-assisted metric that decomposes long captions into QA-style checks.

\section{Challenges and Future Directions}\label{sec:future}
We highlight challenges and promising forward paths that connect SFT, RL, and TTS for video LMMs with a focus on verifiability, efficiency, and robustness. Rather than treating these paradigms as isolated techniques, the field is moving toward deep integration that converts training-time investment into dependable test-time accuracy while addressing concrete limitations reported across recent studies.

\begin{myboxi}[Takeaways]
\begin{itemize}
\item Ground supervision and evaluation in structured, evidence-linked reasoning and explicit verifier signals; actively diagnose and mitigate sycophancy, judge and length biases, and subtitle leakage.
\item Scale RL on long videos with verifiable, compositional rewards, efficient frame selection and caching, and exploration objectives that go beyond distilled teachers.
\item Build budget-aware anytime agents that couple confidence estimates with verifier checks and tool use; standardize reporting (viewing budget, reasoning length, paths, latency/throughput, subtitle usage) to ensure fair comparison and avoid leakage.
\end{itemize}
\end{myboxi}

\subsection{Future Directions for Video-LMM Supervised Fine-Tuning}

\paragraph{Structured interfaces and grounded CoT.}
Codifying reasoning formats that bind steps to evidence (timestamps, frame IDs, regions) can improve faithfulness and simplify verifier design, building on multimodal CoT resources~\cite{han2024videoespressolargescalechainofthoughtdataset, zhang2025vitcotvideotextinterleavedchainofthought, wang2025cotaskschainofthoughtbasedvideo}. Normalizing tags, citations, and unit conventions enables plug-and-play checks later used in RL and TTS.

\paragraph{Verifier-in-the-loop CoT synthesis at scale.}
Automate draft–refine–audit pipelines that start from ASR/OCR/shot metadata, refine on frames, and filter with lightweight checkers to reduce hallucinations. Reduce template and single-model biases by mixing trace generators and including self-correction exemplars; couple instruction tuning to task metrics rather than style alone~\cite{sun2025evaluationdefenseadvancingsafety,lee2025vidchainchainoftasksmetricbaseddirect,chen2025expandingperformanceboundariesopensource}.

\paragraph{Trimodal supervision and subtitle controls.}
Many queries hinge on audio cues and speaker turns. Extend SFT to align speech, events, and visual evidence and always report with and without transcripts to avoid shortcutting via ASR. Current works highlight limited audio coverage and the need for streaming-aware alignment~\cite{sun2025videosalmonno1reasoningenhancedaudiovisuallarge,chen2025scalingrllongvideos,yang2025svbenchbenchmarktemporalmultiturn}.

\paragraph{Hallucination-aware instruction tuning.}
Incorporating counterfactual and absence cases from robustness resources~\cite{gao2025exploringhallucinationlargemultimodal, li2025vidhallucevaluatingtemporalhallucinations} teaches calibrated abstention and verification-seeking behavior, reducing over-affirmation as chains lengthen.

\paragraph{Multilingual, OCR, and narrative structure.}
Data remains imbalanced across languages and misses hard OCR and narrative dependencies. Future SFT should target multilingual breadth, degraded text, and long-span story reasoning so improvements transfer beyond narrow scenarios~\cite{yang2025vidtextcomprehensiveevaluationvideo,zhang2025seriesbenchbenchmarknarrativedrivendrama}.

\subsection{Challenges and Future Directions of RL for Video Reasoning}

\paragraph{Compositional, verifiable rewards.}
Beyond tIoU/IoU, many tasks require joint time–space–semantics checks (entity linking, ordering, object–action binding)~\cite{xu2025viarladaptivetemporalgrounding, luo2025musegreinforcingvideotemporal, li2025videochatr1enhancingspatiotemporalperception}. Process Reward Models (PRMs) can provide dense credit along chains but need cost-effective construction and bias control~\cite{chen2025scalingrllongvideos}. Lightweight rule systems like VeriPO complement PRMs and transfer to TTS verification~\cite{li2025veripocultivatinglongreasoning}.

\paragraph{Sample efficiency and long-video cost.}
Caching visual features and decoupled encoders help~\cite{chen2025scalingrllongvideos}, yet scaling RL still strains budgets. Off-policy and model-based variants, world models, and micro-rollouts (optimize locate-first, then answer) are promising for exploration efficiency~\cite{dubois2025videoeventreasoningprediction}. Architectural context-scaling offers another path. For instance, VideoNSA~\cite{song2025videonsa} applies learnable, hardware-aware native sparse attention~\cite{yuan2025native}, reliably scaling to 128K tokens and improving temporal reasoning over compression-based baselines; MovieChat+~\cite{song2025moviechat+} uses question-aware sparse memory to support long-video reasoning without external temporal modules while cutting cost.

\paragraph{Exploration beyond teachers.}
Curriculum and teacher distillation mitigate cold starts~\cite{chen2025scalingrllongvideos}, but discovering strategies surpassing teachers requires diversity-driven objectives and self-play. Difficulty-aware and group-relative schemes from recent RL for video provide practical starting points~\cite{zhang2025thinkingvideosmultimodaltoolaugmented}.

\paragraph{Evaluation bias and fair comparison.}
Judge bias and length bias can distort progress when using LLMs as evaluators. Report matched budgets, control for reasoning length, and include human or verifier-based audits to ensure reliability~\cite{ma2025videoevalprorobustrealisticlong,zhao2025mmvumeasuringexpertlevelmultidiscipline}.

\paragraph{Scaling beyond preference data.}
Automated pipelines~\cite{wang2025videorftincentivizingvideoreasoning} and self-alignment~\cite{kulkarni2025videosaviselfalignedvideolanguage} reduce annotation dependence but must broaden coverage for causal and counterfactual reasoning and diverse domains~\cite{li2025reinforcementlearningtuningvideollms, zhou2025reinforcedmllmsurveyrlbased}.

\subsection{Video-LMM Test-Time Scaling Future Directions}

\paragraph{Confidence-aware, verifier-guided TTS.}
Stopping rules tied to uncertainty, coupled with verifier checks, can deliver anytime accuracy: deepen reasoning or densify viewing only when needed, echoing closed-loop designs and sparse-to-dense schedules~\cite{meng2025cybervcyberneticstesttimescaling, wang2025videortsrethinkingreinforcementlearning}.

\paragraph{Tool-augmented inference and distillation.}
Reasoning that interleaves tool calls (retrieval, tracking, ASR alignment) improves faithfulness at test time~\cite{zhang2025thinkingvideosmultimodaltoolaugmented}; post-hoc distillation can transfer these benefits into base models to cut inference cost, using verifier-anchored traces as supervision~\cite{li2025veripocultivatinglongreasoning}.

\paragraph{Streaming agents with memory.}
Agentic planners that decide what to watch next and when to stop, while maintaining task-aware working memory, are essential for long or streaming video~\cite{xiong2025streamingvideounderstandingmultiround, zhi2025videoagent2enhancingllmbasedagent, wang2025videochata1thinkinglongvideos}. Budget-aware rewards can train these behaviors for robust anytime performance.

\paragraph{Standardized reporting and leakage control.}
Report viewing budgets, reasoning lengths, path counts, latency/throughput, and subtitle usage. Include sycophancy and judge-bias diagnostics so gains are attributable and not artifacts of prompt length or transcript leakage~\cite{zhao2025mmvumeasuringexpertlevelmultidiscipline,chen2025expandingperformanceboundariesopensource}.

\paragraph{Compute–accuracy trade-offs under constrained viewing.}
Co-tune frame selection and compression with reasoning quality so systems remain strong when only a small fraction of frames are processed. Frame-optimization and compression frameworks still incur notable cost; future work should make these components data- and compute-efficient~\cite{lee2025refocusreinforcementguidedframeoptimization,wang2025lvclightweightcompressionframework}.

\section{Conclusion}\label{sec:conclusion}

This survey has systematically analyzed the critical role of post-training in advancing video reasoning, tracing the evolution from foundational Supervised Fine-tuning with Chain-of-Thought to more powerful and autonomous paradigms. Reinforcement learning, primarily through online frameworks like GRPO, has become a core engine for optimization, while emerging agentic frameworks and test-time scaling strategies offer new frontiers in reasoning capability and efficiency. Despite these significant advances, the path to robust, general-purpose video intelligence is still marked by key challenges. The future research agenda will be defined by overcoming data scarcity for complex reasoning, developing more sample-efficient and stable RL algorithms, strengthening multimodal grounding to prevent hallucinations, and creating integrated frameworks that synergize training-time alignment with inference-time computation. Addressing these interconnected challenges is crucial to advancing the boundaries of video understanding systems.



\bibliography{reference}

\end{document}